\documentclass[journal]{IEEEtran}
\usepackage{amsmath,amsopn,amssymb,amsthm}
\usepackage[dvips]{graphicx}
\usepackage{xspace,color,soul}
\usepackage{longtable,multirow,bigdelim}
\usepackage{array,float}
\usepackage{subfigure}
\usepackage{soul}
\usepackage{cite}
\newtheorem{theorem}{Theorem}

\usepackage{algorithm,algorithmicx,algpseudocode}
\usepackage{bm}
\usepackage{url}
\usepackage{breqn}

\newcommand{\blue}[1] {\textcolor[rgb]{0.0,0.0,0.0}{{#1}}}

\begin{document}
\title{3D Point Cloud Denoising Using Graph Laplacian\\Regularization of a Low Dimensional\\Manifold Model}

\author{Jin Zeng,~\IEEEmembership{Member,~IEEE,} Gene Cheung,~\IEEEmembership{Senior Member,~IEEE,} Michael Ng,~\IEEEmembership{Senior Member,~IEEE,}\\ Jiahao Pang,~\IEEEmembership{Member,~IEEE,} and Cheng Yang,~\IEEEmembership{Member,~IEEE}

\thanks{Jin Zeng is with Hong Kong University of Science and Technology, Clear Water Bay, Kowloon, Hong Kong. Email: jzengab@connect.ust.hk

Gene Cheung and Cheng Yang are with the Department of Electrical Engineering \& Computer Science, York University, Toronto, Canada. Email: genec@yorku.ca; cyang@eecs.yorku.ca

Michael Ng is with the Centre for Mathematical Imaging
and Vision, Department of Mathematics, Hong Kong Baptist University,
Hong Kong. Email: mng@math.hkbu.edu.hk

Jiahao Pang is with SenseTime Research, Hong Kong. Email: pangjiahao@sensetime.com
}
}

\maketitle % Print the title

\begin{abstract}
3D point cloud---a new signal representation of volumetric objects---is a discrete collection of triples marking exterior object surface locations in 3D space.  
Conventional imperfect acquisition processes of 3D point cloud---e.g., stereo-matching from multiple viewpoint images or depth data acquired directly from active light sensors---imply non-negligible noise in the data.
In this paper, we extend a previously proposed low-dimensional manifold model for the image patches to surface patches in the point cloud, and seek self-similar patches to denoise them simultaneously using the patch manifold prior. 
Due to discrete observations of the patches on the manifold, we approximate the manifold dimension computation defined in the continuous domain with a patch-based graph Laplacian regularizer, and propose a new discrete patch distance measure to quantify the similarity between two same-sized surface patches for graph construction that is robust to noise. 
We show that our graph Laplacian regularizer leads to speedy implementation and has desirable numerical stability properties given its natural graph spectral interpretation.
Extensive simulation results show that our proposed denoising scheme outperforms state-of-the-art methods in objective metrics and better preserves visually salient structural features like edges.
\end{abstract}


\section{Introduction}
\label{sec:intro}
The three-dimensional (3D) point cloud has become an important and popular signal representation of volumetric objects in 3D space \cite{rusu20113d,thanou2016graph,chen2018fast}. 
3D point cloud can be acquired directly using low-cost depth sensors like Microsoft Kinect or high-resolution 3D scanners like LiDAR.  
Moreover, multi-view stereo-matching techniques have been extensively studied in recent years to recover a 3D model from images or videos, where the typical output format is the point cloud \cite{ji2017surfacenet}. However, in either case, the output point cloud is inherently noisy, which has led to numerous approaches for point cloud denoising \cite{rosman2013patch,matteipoint,sun2015denoising,zheng2017guided}. 

Moving least squares (MLS)-based \cite{guennebaud2007algebraic,oztireli2009feature} and locally optimal projection (LOP)-based methods \cite{lipman2007parameterization,huang2013edge} are two major categories of point cloud denoising approaches, but are often criticized for over-smoothing \cite{sun2015denoising, zheng2017guided} due to the use of local operators.
Sparsity-based methods, based on the local planarity assumption, are optimized towards a sparse representation of certain geometric features such as surface normals \cite{avron2010ℓ,sun2015denoising} and
point deviations from local reference plane \cite{matteipoint}.
They were reported to provide the state-of-the-art performance \cite{han2017review}. However at high noise levels, the inaccurate estimation for normal or the local plane can lead to over-smoothing or over-sharpening \cite{sun2015denoising,matteipoint}.

Non-local methods generalize the non-local means \cite{buades2005non} and BM3D \cite{dabov2007image} image denoising algorithms to point cloud denoising, and are shown to better preserve fine shape features under high level of noise.
The approaches in \cite{wang2008similarity, deschaud2010point} extend the non-local means denoising approach to point clouds and adaptively filter the points in an edge preserving manner.
% A more recent method in \cite{rosman2013patch} is inspired by BM3D and exploits the inherent self-similarity between surface patches to preserve structural details, but the computational complexity is too high to be practical.
\cite{rosman2013patch} is inspired by BM3D and exploits the inherent self-similarity between surface patches to preserve structural details, but the computational complexity is too high to be practical.
\blue{A more recent method in \cite{sarkar2018structured} also utilizes the patch self-similarity and denoises the local patches based on dictionary learning.}

Utilizing an assumed self-similarity characteristic in images has long been a popular strategy in image processing \cite{buades2005non,dabov2007image}. 
Extending on these earlier works, a more recent work \cite{osher2017low} proposed the \textit{low-dimensional manifold model} (LDMM) for image processing, assuming that similar image patches are samples of a low-dimensional manifold in high-dimensional space. 
The assumption is verified in various applications in image processing and computer vision \cite{peyre2009manifold,peyre2011review}.
In LDMM, the manifold dimension is used for regularization to recover the image, achieving state-of-the-art results in various inverse imaging applications, \textit{e.g.}, denoising, inpainting, superresolution, \textit{etc.}.

Inspired by the LDMM work in \cite{osher2017low}, we exploit self-similarity of the surface patches by assuming that the surface patches in the point cloud lie on a manifold of low dimension. 
However, the extension of LDMM from images to point clouds is non-trivial. 
First, the computation of manifold dimension requires a well-defined coordinate function in \cite{osher2017low}, \textit{i.e.}, the extrinsic coordinates of points on the manifold, which is straightforward for image patches but not for surface patches due to the irregular structure of point clouds. 
Moreover, the point integral method (PIM) for solving the dimension optimization in \cite{osher2017low} is of high complexity. In the outer loop, the manifold and the image are iteratively updated, while in the inner loop, the coordinate function and pixel values are updated until convergence. Since the linear systems for updating coordinate function are asymmetric due to the constraints enforced by PIM, a large number of iterations is required to reach convergence, leading to high computational cost \cite{shi2018generalization}.

To address the two issues above, we approximate the patch-manifold dimension defined in continuous domain with a discrete patch-based graph Laplacian regularizer (GLR). 
%due to discrete observation of the surface patches. 
Specifically, the main contributions of our work are as follows:
\begin{enumerate}
\item By adopting the LDMM, we exploit the surface self-similarity characteristic and simultaneously denoise similar patches to better preserve sharp features;
\item By approximating the computation of the manifold dimension with GLR, we avoid explicitly defining the manifold coordinate functions and enable the LDMM to extend to the point cloud setting; 
\item By using GLR, the implementation is accelerated with a reduced number of iterations thanks to the symmetric structure of the graph Laplacian matrix;
\item Our GLR is shown to provide a graph spectral interpretation and is guaranteed numerical stability via eigen-analysis in the graph spectral domain \cite{shuman13};
\item An efficient similarity measure for discrete $k$-pixel patch pairs is designed for graph construction that is robust to noise.
\end{enumerate}
Extensive simulation results show that our proposed method outperforms the state-of-the-art methods in objective metrics and better preserves visually salient features like edges.

The rest of the paper is organized as follows. Section\;\ref{sec:related} overviews some existing works. Section\;\ref{sec:ld} defines the patch manifold associated with the 3D point cloud. Section\;\ref{sec:denoising} formulates the denoising problem by describing how the manifold dimension is computed and approximated with the graph Laplacian regularizer. The algorithm implementation is discussed in Section\;\ref{sec:algo} with graph spectral analysis to interpret the algorithm and a numerical stability analysis. 
Finally, Section\;\ref{sec:result} and Section\;\ref{sec:con} presents experimental results and concludes the paper respectively.

\section{Related Work}
\label{sec:related}
Previous point cloud denoising works can be classified into four categories: moving least squares (MLS)-based methods, locally optimal projection (LOP)-based methods, sparsity-based methods, and non-local similarity-based methods.

\textbf{MLS-based methods.} MLS-based methods approximate a smooth surface from the input samples and project the points to the resulting surface. To construct the surface, the method in \cite{alexa2003computing} first finds the local reference domain for each point that best fits its neighboring points in terms of MLS, then defines a function based on the reference domain by fitting a polynomial function to neighboring data. 

Several extensions, which address the unstable reconstruction problem in the case of high curvature, \textit{e.g.}, algebraic point set surfaces (APSS) \cite{guennebaud2007algebraic} and its variant in \cite{guennebaud2008dynamic}, or preserve the shape features, \textit{e.g.}, robust MLS (RMLS) \cite{rusu2007towards} and robust implicit MLS (RIMLS) \cite{oztireli2009feature}, have also been proposed. These methods can robustly generate a smooth surface from extremely noisy input, but are often criticized for over-smoothing \cite{sun2015denoising, zheng2017guided}.

\textbf{LOP-based methods.} Unlike MLS-based methods, LOP-based methods do not compute explicit parameters for the surface. For example, LOP method in \cite{lipman2007parameterization} outputs a set of points that represent the underlying surface while enforcing a uniform distribution over the point cloud with a repulse term in the optimization. 
Its modifications include weighted LOP (WLOP) \cite{huang2009consolidation}, which provides a more uniformly distributed output by adapting the repulse term to the local density, and anisotropic WLOP (AWLOP) \cite{huang2013edge}, which preserves sharp features by modifying WLOP to use an anisotropic weighting function. LOP-based methods also suffer from over-smoothing due to the use of local operators, or generate extra features caused by noise \cite{sun2015denoising, zheng2017guided}.

\textbf{Sparsity-based methods.} Sparsity-based methods are based on a local planarity assumption and optimize for sparse representations of certain geometric features. Methods based on the sparsity of surface normals would first obtain a sparse reconstruction of the surface normals by solving a global minimization problem with $l_1$ \cite{avron2010ℓ} or $l_0$ \cite{sun2015denoising} regularization, then update the point positions with the surface normals by solving another global minimization problem based on the locally planar assumption. 
A more recent method called Moving Robust Principal Components Analysis (MRPCA) \cite{matteipoint} uses $l_1$ minimization of the point deviations from the local reference plane to preserve sharp features.
Sparsity-based approaches are reported to achieve the state-of-the-art performance \cite{han2017review}, though at a high level of noise, the estimation of normal or local plane can be so poor that it leads to over-smoothing or over-sharpening \cite{sun2015denoising}.

\textbf{Non-local methods.} \blue{Non-local methods are widely adopted in image denoising \cite{zha2018rank,zha2018group,zha2017image,zha2017imaged,wang2017nonconvex}.} Non-local methods generalize the notion of non-local self-similarity in the non-local means \cite{buades2005non} and BM3D \cite{dabov2007image} image denoising algorithms to point cloud denoising, and are shown to better preserve structural features under high level of noise. 

Due to the lack of regular structure in a point cloud, extending non-local image denoising schemes to point cloud is difficult.
\cite{wang2008similarity} utilizes curvature-based similarity to perform non-local filtering, so that the filtering considers the neighborhood geometry structure and better preserves fine shape features.
\cite{deschaud2010point} proposes to use the polynomial coefficients of the local MLS surface as neighborhood descriptors to compute point similarity.
% Some early trials are \cite{guillemot2012non} and \cite{digne2012similarity} \red{to change}, which utilize a non-local means algorithm.

Inspired by the BM3D algorithm, \cite{rosman2013patch} exploits self-similarity among surface patches in the point cloud and outperformes the non-local means methods. However, the computational complexity is typically too high to be practical, taking a few hours for a point cloud of size 15,000 as reported in \cite{rosman2013patch}.
\blue{A more recent method in \cite{sarkar2018structured} also utilizes patch self-similarity and optimizes for a low-rank dictionary representation of the extracted patches to impose patch smoothness. During patch extraction, the points in each patch are projected to a regular grid for subsequent linear operations where multiple points can fall to the same location, leading to lose of fine structure and over-smoothing. The method is referred to as LR for short hereinafter.} 

Our method belongs to the fourth category, the non-local methods. Similar to \cite{rosman2013patch,sarkar2018structured}, we also utilize the self-similarity among patches via the low-dimensional manifold prior \cite{osher2017low}. 
However, the original PIM for manifold dimension minimization in \cite{osher2017low} is not applicable to the point cloud setting due to the lack of regular structure of surface patches to define coordinate functions. 
Even if the coordinate functions are provided, PIM is time-consuming because the linear systems derived from PIM are asymmetric and inefficient to solve. 
In contrast, thanks to GLR, our approach eliminates the need for coordinate functions and can be efficiently implemented, outperforming existing schemes with better feature preservation.

In \cite{shi2018generalization}, PIM is approximated with the weighted nonlocal graph Laplacian (WNLL) to reduce computational complexity. The WNLL also preserves the symmetry of the linear systems with a graph Laplacian to speed up the implementation, but the Laplacian matrix is derived from the Laplace-Beltrami equation in PIM thus \textit{different} from our GLR. 
Nevertheless, similar to PIM, the WNLL approach is designed for image restoration and solves each coordinate function separately, thus cannot be directly applicable to point clouds.
%\red{GC: as stated in the response letter, [20] is pixel -based, while ours is patch-based, and thus the methods are fundamentally different and ours cannot be interpreted as an extension of [20]. Is that correct? if so, we can repeat our argument in the response letter here.}

% \section{Preliminaries}
% \label{sec:pre}
% \input{preliminary}

\section{Patch Manifold}
\label{sec:ld}
We first define the notion of \textit{patch manifold} given a point cloud $\mathcal{V} = \{\mathbf{v}_i\}_{i=1}^N$, $\mathbf{v}_i \in \mathbb{R}^3$, which is a (roughly uniform) discrete sampling of a 2D surface of a 3D object. 
%\blue{(Here I add matrix representation at the beginning)} 
Let $\mathbf{V} = [\mathbf{v}_1,\dots,\mathbf{v}_N]^{\top}\in \mathbb{R}^{N \times 3}$ be the position matrix for the point cloud.
Noise-corrupted $\mathbf{V}$ can be simply modeled as: 
\begin{equation}
\mathbf{V} = \mathbf{U} + \mathbf{E},
\end{equation}
where $\mathbf{U}$ contains the true 3D positions, $\mathbf{E}$ is a zero-mean signal-independent noise (we assume Gaussian noise in our experiments), and $\mathbf{U}, \mathbf{E} \in \mathbb{R}^{N \times 3}$. 
%\red{gene: any assumption on noise? Not sure here, the derivation doesn't assume specific model for the noise, but would usually assume zero-mean I suppose.} 
To recover the true position $\mathbf{U}$, we consider the \textit{low-dimensional manifold model} prior (LDMM) \cite{osher2017low} as a regularization term for this ill-posed problem.

\subsection{Surface Patch}

We first define a \textit{surface patch} in a point cloud. 
We select a subset of $M$ points from $\mathcal{V}$ as the \textit{patch centers}, \textit{i.e.}, $\{\mathbf{c}_m\}_{m=1}^{M} \subset \mathcal{V}$. 
Then, patch $p_m$ centered at a given center $\mathbf{c}_m$ is defined as the set of $k$ nearest neighbors of $\mathbf{c}_m$ in $\mathcal{V}$, in terms of Euclidean distance. 

The union of the patches should cover the whole point cloud, \textit{i.e.}, $\bigcup_{m=1}^{M} p_m = \mathcal{V}$. There can be different choices of patch centers, and the degree of freedom can be used to trade off computation cost and denoising performance. 
Let $\mathbf{p}_m \in \mathbb{R}^{3k}$ be the patch coordinates, composed of the $k$ points in $p_m$. 
%\red{gene: should the dimension of $\mathbf{p}_m$ be $\mathbb{R}^{k \times 3}$ instead, to be consistent with earlier definitions?}

\subsection{Patch Manifold}

Here we adopt the basic assumption in \cite{osher2017low} that the patches sample a low-dimensional smooth manifold embedded in $\mathbb{R}^{3k}$, which is called the \textit{patch manifold} $\mathcal{M}(\mathbf{U})$ associated with the point cloud $\mathbf{U}$. 
In order to evaluate similarity among patches, we first need to align the patches; \textit{i.e.}, the coordinates $\mathbf{p}_m$ should be translated with respect to $\mathbf{c}_m$, so that $\mathbf{c}_m$ lies on the origin $(0,0,0)$. Hereafter we set $\{\mathbf{p}_m\}_{m=1}^{M}$ to be the translated coordinates.

\subsection{Low Dimensional Patch Manifold Prior}

The LDMM prior assumes that the solution contains patches that minimize the patch manifold dimension. 
We can thus formulate a \textit{maximum a posteriori} (MAP) problem with prior and fidelity terms as follows:
%\red{gene:u defined $M$ patches $p_M$ each with $k$ points earlier, but here actually the objective does not involved how the patches are defined, but only the patch manifold $\mathcal{M}$, which is a function of the denoised points $\mathbf{U}$ only. Is this ok?} \blue{explained later}
\begin{equation}
\underset{\mathbf{U}}{\text{min}} \quad \text{dim}(\mathcal{M}(\mathbf{U}))
+ \lambda \| \mathbf{V} - \mathbf{U} \|_F^2,
\label{eq:obj0}
\end{equation}
where $\lambda$ is a parameter that trades off the prior with the fidelity term, and 
$\|.\|_F^2$ is the Frobenius norm. 
Note that given a certain strategy of patch selection,
the patches are determined by the point cloud $\mathbf{U}$, and the patches in turn define the underlying manifold $\mathcal{M}$. 
Hence we view $\mathcal{M}$ as a function of $\mathbf{U}$.

The patches can be very different and sampled from different manifolds of different dimensions.
For example, a flat planar patch belongs to a manifold of lower dimension than a patch with corners.
The dimension of the patch manifold, $\text{dim}(\mathcal{M}(\mathbf{U}))$ becomes a function of the patch, and the integration of $\text{dim}(\mathcal{M}(\mathbf{U}))$ over $\mathcal{M}$ is used as the regularization term,
%\red{so for different $\mathbf{p}$ on $\mathcal{M}$ u get different dimensions, and by integrating over all $\mathbf{p}$ u r taking some kind of average?} \blue{yes, same in \cite{osher2017low}}
\begin{equation}
\underset{\mathbf{U}}{\text{min}} \quad \int_{\mathcal{M}} \text{dim}(\mathcal{M}(\mathbf{U}))(\mathbf{p})d \mathbf{p} + \lambda \| \mathbf{V} - \mathbf{U} \|_F^2,
\label{eq:obj0b}
\end{equation}
where $\text{dim}(\mathcal{M}(\mathbf{U}))(\mathbf{p})$ is the dimension of $\mathcal{M}(\mathbf{U})$ at $\mathbf{p}$.
Here $\mathbf{p} \in \mathbb{R}^{3k}$ is a point on $\mathcal{M}$.
The question that remains is how to compute $\text{dim}(\mathcal{M}(\mathbf{U}))(\mathbf{p})$. In the next section, the dimension computation is mathematically defined and approximated with GLR.

\section{Problem Formulation}
\label{sec:denoising}
In this section, we first briefly review the calculation of the manifold dimension in continuous domain, then approximate this computation with the GLR so as to efficiently adopt LDMM to discrete point cloud patches. 
%Moved to next section: Moreover, we design a patch distance measure for the graph construction on the manifold.

\subsection{Manifold Dimension Computation in Continuous Domain}

Here we overview how the manifold dimension is computed in \cite{osher2017low}.
First, let $\alpha_i$, where $i=1,\ldots,3k$, be the coordinate functions on the manifold $\mathcal{M}$ embedded in $\mathbb{R}^{3k}$, \textit{i.e.},
\begin{equation} \label{eq:alpha}
\alpha_i(\mathbf{p}) = p_i, ~~ \forall \mathbf{p}=[p_1,\dots,p_{3k}]^{\top} \in \mathcal{M}. 
\end{equation}

According to \cite{osher2017low}, the dimension of $\mathcal{M}$ at $\mathbf{p}$ is given by:
\begin{equation} 
\text{dim}(\mathcal{M})(\mathbf{p}) = \sum_{i=1}^{3k} \|\nabla_{\mathcal{M}} \alpha_i(\mathbf{p}) \|^2,
\end{equation}
where $\nabla_{\mathcal{M}} \alpha_i(\mathbf{p})$ denotes the gradient of the function $\alpha_i$ on $\mathcal{M}$ at $\mathbf{p}$.
Then the integration of $\mathrm{dim}(\mathcal{M})(\mathbf{p})$ over $\mathcal{M}$ is given as,
\begin{equation} \label{eq:osher}
\int_{\mathcal{M}} \text{dim}(\mathcal{M})(\mathbf{p}) d \mathbf{p} = \sum_{i=1}^{3k} \int_{\mathcal{M}} ||\nabla_{\mathcal{M}} \alpha_i(\mathbf{p})||^2 d \mathbf{p}.
\end{equation}
%\red{u mean u r integrating over all samples $\mathbf{p}$ on manifold $\mathcal{M}$, right?}
%\blue{(\ref{eq:osher}) follows from (\ref{eq:obj0b}), it's the integral over the manifold $\mathcal{M}$.}

The formula in (\ref{eq:osher}) is a sum of integrals on continuous manifold $\mathcal{M}$ along different dimensions, but our observations $\{\mathbf{p}_m\}_{m=1}^M$ of the manifold $\mathcal{M}(\mathbf{U})$ are discrete and finite.
In \cite{osher2017low}, the solution to the dimension minimization is given by a partial derivative equation (PDE) for each $\alpha_i$ separately, which is discretized at the patch observations using PIM, solved via a linear system.  
% leading to a Laplacian-Beltrami operator. 
% The PIM is originally designed for image patches, while in the setting of patch-based point cloud denoising, the PIM is not applicable because the definition of coordinate function is difficult.

However, PIM requires the patch coordinates $\{\mathbf{p}_m\}_{m=1}^M$ to be ordered so that the $\boldsymbol{\alpha}_i$'s can be defined. 
For example, if the patches are image patches of the same size, then the patch coordinates are naturally ordered according to pixel location, \textit{i.e.}, the $i$-th entry in $\mathbf{p}_m$ is the pixel value at the $i$-th location in the image patch. 
However, surface patches in the 3D point cloud are unstructured, and there is no natural way to implement global coordinate ordering for all patches.

This motivates us to discretize the manifold dimension with GLR, eliminating the need for global ordering and can be implemented efficiently.

\subsection{Dimension Discretization with GLR}
% \subsection{Discretization of Manifold Dimension}

We first introduce the graph construction on a manifold, which induces the GLR. 
Then we discuss how the GLR approximates the manifold dimension and avoids global coordinate ordering.

\subsubsection{Constructing Graph on a Manifold}

We construct a discrete graph $\mathcal{G}$ whose vertex set is the observed surface patches $\mathcal{P} = \{\mathbf{p}_m\}_{m=1}^M$ lying on $\mathcal{M}(\mathbf{U})$, \textit{i.e.}, $\mathbf{p}_m \in \mathcal{M}(\mathbf{U}) \subset \mathbb{R}^{3k}$.
Let $\mathcal{E}$ denote the edge set, where the edge between $m$-th and $n$-th patches is weighted as,
\begin{equation} \label{eq:w}
w_{mn} = (\rho_m \rho_n)^{-1/\gamma} \psi(d_{mn}).
\end{equation}
The kernel $\psi(\cdot)$ is a thresholded Gaussian function
\begin{equation} \label{eq:kernelfunc}
\psi(d_{mn})=
\begin{cases}
\exp(-\frac{d_{mn}^2}{2\epsilon^2}) & d_{mn}<r \\
0& \text{otherwise}, 
\end{cases} 
\end{equation}
and $d_{mn}$ is the Euclidean distance between the two patches $\mathbf{p}_m$ and $\mathbf{p}_n$,
\begin{equation}\label{eq:dmn}
d_{mn} = ||\mathbf{p}_m - \mathbf{p}_n||_2.
\end{equation}
The term $(\rho_m \rho_n)^{-1/\gamma}$ is the normalization term, where $\rho_n = \Sigma_{m=1}^M \psi(d_{mn})$ is the degree of $\mathbf{p}_n$ before normalization. 
The graph constructed in these settings is an $r$-neighborhood graph, \textit{i.e.}, no edge has a distance greater than $r$. Here $r = \epsilon C_r$, and $C_r$ is a constant.

\subsubsection{Graph Laplacian Regularizer}

With the edge weights defined above, we define the symmetric adjacency matrix $\mathbf{A} \in \mathbb{R}^{M \times M}$, with the $(m,n)$-th entry given by $w_{mn}$. $\mathbf{D}$ denotes the diagonal degree matrix, where entry $\mathbf{D}(m,m) = \sum_n w_{m,n}$. The combinatorial graph Laplacian matrix is $\mathbf{L} = \mathbf{D} - \mathbf{A}$ \cite{shuman13}. 

For the coordinate function $\alpha_i$ on $\mathcal{M}$ defined in (\ref{eq:alpha}), sampling $\alpha_i$ at positions of $\mathcal{P}$ leads to its discretized version, $\boldsymbol{\alpha}_i = [\alpha_i(\mathbf{p}_1) \dots \alpha_i(\mathbf{p}_M)]^\top$. The graph Laplacian $\mathbf{L}$ induces the regularizer $S_{\mathbf{L}}(\boldsymbol{\alpha}_i) = \boldsymbol{\alpha}_i^{\top} \mathbf{L} \boldsymbol{\alpha}_i$. It can be shown that
\begin{equation}
S_{\mathbf{L}}(\boldsymbol{\alpha}_i) = \boldsymbol{\alpha}_i^{\top} \mathbf{L} \boldsymbol{\alpha}_i =  \sum_{(m,n)\in \mathcal{E}} w_{mn} (\alpha_i(\mathbf{p}_m) - \alpha_i(\mathbf{p}_n))^2.
\end{equation}

\subsubsection{Approximation with Graph Laplacian Regularizer}
\label{sec:normalized}
We now show the convergence of the discrete graph Laplacian regularizer to the dimension of the underlying continuous manifold. 

First, we declare the following theorem that relates $S_{\mathbf{L}}(\boldsymbol{\alpha}_i)$ to the integral of $\|\nabla_{\mathcal{M}} \alpha_i(\mathbf{p})\|_2^2$ on $\mathcal{M}$ on the right side of (\ref{eq:osher}):
\begin{theorem} \label{theorem:1}
Under conditions specified in Appendix \ref{sec:append} for $\epsilon$, $\mathcal{M}$ and function $\boldsymbol{\alpha}_i$,
\begin{equation} \label{eq:approx0}
\underset{\substack{M \rightarrow \infty,\\ \epsilon \rightarrow 0, \delta \rightarrow 0}} {\lim}  S_{\mathbf{L}}(\boldsymbol{\alpha}_i)
\sim \frac{1}{|\mathcal{M}|} \int_{\mathcal{M}} \|\nabla_{\mathcal{M}} \alpha_i(\mathbf{p})\|_2^2 d \mathbf{p}, 
\end{equation}
where $|\mathcal{M}|$ is the volume of the manifold $\mathcal{M}$, $\delta$ is the manifold dimension, and $\sim$ means there exists a constant depending on $\mathcal{M}$, $C_r$ and $\gamma$, such that the equality holds.
\end{theorem}
In other words, as the number of samples $M$ increases and the neighborhood size $r=\epsilon C_r$ shrinks, $S_{\mathbf{L}}(\boldsymbol{\alpha}_i)$ approaches its continuous limit.
Moreover, if the manifold dimension $\delta$ is low, we can ensure a good approximation of the continuous regularization functional even if the manifold is embedded in a high-dimensional space. Detailed proof for the above theorem is provided in Appendix \ref{sec:append}.

Consequently, given a point cloud, one can approximate the dimension of $\mathcal{M}$ with the $\boldsymbol{\alpha}_i$'s and the constructed graph Laplacian $\mathbf{L}$ following (\ref{eq:osher}) and (\ref{eq:approx0}):
\begin{align} \nonumber
    \underset{\substack{M \rightarrow \infty,\\ \epsilon \rightarrow 0, \delta \rightarrow 0}} {\lim} |\mathcal{M}| \sum_{i=1}^{3k} \boldsymbol{\alpha}_i^\top \mathbf{L} \boldsymbol{\alpha}_i
&\sim \sum_{i=1}^{3k} \int_{\mathcal{M}} \|\nabla_{\mathcal{M}} \alpha_i(\mathbf{p})\|_2^2 d \mathbf{p}\\ \label{eq:approx}
&=\int_{\mathcal{M}} \text{dim}(\mathcal{M})(\mathbf{p}) d \mathbf{p}.
\end{align}
%\red{gene: but how do u order the $k$ points in each patch so that the $i$-th point in patch 1 corresponds to the $i$-th point in patch 2?} 
%\blue{I didn't do the ordering globally, like patch 1 to M are all in order. Instead the ordering is done locally, for connected patch pair, they are ordered pairwisely. This is explained as follows.}

\blue{Note that Theorem \ref{theorem:1} is derived based on the combinatorial Laplacian matrix and does not apply to other types of Laplacian, \textit{e.g.}, normalized Laplacian $\mathcal{L} = \mathbf{D}^{-1/2}\mathbf{L}\mathbf{D}^{-1/2}$. Further, a regularizer using $\mathcal{L}$ would penalize a constant signal, since the eigenvector corresponding to eigenvalue $0$ is not constant \cite{liu2017random}, which means it cannot handle constant signal. Experimental comparison between combinatorial and normalized Laplacian is provided in Section \ref{sec:iters}.}

\subsection{From Global Coordinate Ordering to Local Correspondence}
So far, we obtain the approximation in (\ref{eq:approx}), 
but the above graph construction still requires the patch coordinates $\{\mathbf{p}_m\}_{m=1}^M$ to be ordered so that the $\boldsymbol{\alpha}_i$'s can be defined and the patch distance $d_{mn}$ in (\ref{eq:dmn}) determines the patch similarity. 

% \blue{For example, if the patches are image patches of the same size, then the patch coordinates are naturally ordered according to pixel location, \textit{i.e.}, the $i$-th entry in $\mathbf{p}_m$ is the pixel value at the $i$-th location in the image patch. 
% However, surface patches in the 3D point cloud are unstructured, and there is no natural way to implement global coordinate ordering for all patches. (move to front) }

In the following, we argue that the computation of the regularization term can be accomplished based on local pairwise correspondence between connected patches, relieving the need for global ordering. 

We modify the manifold dimension formula in (\ref{eq:approx}):
\begin{align} \label{eq:local1}
\sum_{i=1}^{3k} \boldsymbol{\alpha}_i^\top \mathbf{L} \boldsymbol{\alpha}_i 
&= \sum_{i=1}^{3k} \sum_{(m,n)\in \mathcal{E}} w_{mn} (\alpha_i(\mathbf{p}_m) - \alpha_i(\mathbf{p}_n))^2 \\  \label{eq:local2}
&= \sum_{(m,n)\in \mathcal{E}} w_{mn} \sum_{i=1}^{3k} (\alpha_i(\mathbf{p}_m) - \alpha_i(\mathbf{p}_n))^2 \\  \label{eq:local3} 
&= \sum_{(m,n)\in \mathcal{E}} w_{mn} d_{mn}^2,
\end{align}
where (\ref{eq:local3}) follows from (\ref{eq:local2}) according to the definition of $d_{mn}$ in (\ref{eq:dmn}).
From (\ref{eq:local3}) we see that $\boldsymbol{\alpha}_i$ is not necessary to compute the graph Laplacian regularizer, and hence global coordinate ordering is not required. 
Moreover, since $w_{mn}$ is itself a function of $d_{mn}$ via (\ref{eq:w}), we can obtain the manifold dimension as long as $d_{mn}$ is given by finding the local pairwise correspondence between neighboring patches. 

To reformulate (\ref{eq:local3}) into matrix form, we first consider the subgraphs composed of connected patch pair to reformulate $w_{mn} d^2_{mn}$, then sum up the weights between patch pairs to give the final GLR.

For a connected patch pair $\mathbf{p}_m$ and $\mathbf{p}_n$, let $\mathbf{p}^{\Theta}_{mn} = \begin{bmatrix}\mathbf{p}_{m,\Theta}^\top & \mathbf{p}_{n,\Theta}^\top \end{bmatrix}^\top$ be the concatenation of $\mathbf{p}_m$ and $\mathbf{p}_n$ coordinates, where $\Theta \in \{x,y,z \}$ denotes the 3D coordinates. 
Given the local correspondence between $\mathbf{p}_m$ and $\mathbf{p}_n$, we connect corresponding points to construct the subgraph and multiply the edge weights with $w_{mn}$, resulting in the graph Laplacian matrix $\mathbf{L}_{mn}$ for this subgraph. $w_{mn} d^2_{mn}$ is then reformulated as:
\begin{equation} 
w_{mn} d^2_{mn} 
= \sum_{\Theta \in \{x,y,z\}} (\mathbf{p}^{\Theta}_{mn})^{\top} \mathbf{L}_{mn} \mathbf{p}^{\Theta}_{mn}. 
\end{equation}
Let $\mathbf{S}_{mn} \in \{0, 1\}^{2k \times kM} $ be the sampling matrix to extract $\mathbf{p}^{\Theta}_{mn}$ from $\mathbf{P}_{\Theta}$, where $\mathbf{P}_{\Theta}$ is the coordinate vector of points in all patches, \textit{i.e.}, $\mathbf{p}^{\Theta}_{mn} = \mathbf{S}_{mn} \mathbf{P}_{\Theta}$, so that $w_{mn} d^2_{mn}$ becomes:
\begin{align} 
w_{mn} d^2_{mn}  
&=  \sum_{\Theta \in \{x,y,z\}} (\mathbf{S}_{mn} \mathbf{P}_{\Theta})^{\top} \mathbf{L}_{mn} (\mathbf{S}_{mn}\mathbf{P}_{\Theta}) . 
\end{align}
Then the manifold dimension becomes:
\begin{align} 
&\sum_{(m,n)\in \mathcal{E}} w_{mn} d_{mn}^2  \\
=&\sum_{\Theta \in \{x,y,z\}} \sum_{(m,n)\in \mathcal{E}} \mathbf{P}_{\Theta}^{\top} (\mathbf{S}_{mn}^{\top} \mathbf{L}_{mn} \mathbf{S}_{mn}) \mathbf{P}_{\Theta}\\
=&\sum_{\Theta \in \{x,y,z\}} \mathbf{P}_{\Theta}^{\top} \big( \sum_{(m,n)\in \mathcal{E}} \mathbf{S}_{mn}^{\top} \mathbf{L}_{mn} \mathbf{S}_{mn} \big) \mathbf{P}_{\Theta}\\
=&\sum_{\Theta \in \{x,y,z\}} \mathbf{P}_{\Theta}^{\top} \mathbf{L}_p \mathbf{P}_{\Theta},
\end{align}
where 
\begin{equation}\label{eq:Lp}
    \mathbf{L}_p = \sum_{(m,n)\in \mathcal{E}} \mathbf{S}_{mn}^{\top} \mathbf{L}_{mn} \mathbf{S}_{mn} \in \mathbb{R}^{kM \times kM}
\end{equation}
is the overall graph Laplacian matrix for the point-domain graph.

\subsection{Objective Formulation with GLR Prior}

With $\mathbf{L}_p$ calculated as described above, the optimization is reformulated as:
\begin{equation} \label{eq:mu}
\underset{\mathbf{U}}{\text{min}} \quad \sum_{\Theta \in \{x,y,z\}} \mathbf{P}_{\Theta}^{\top} \mathbf{L}_p \mathbf{P}_{\Theta} + \mu \left\| \mathbf{V} - \mathbf{U}\right\|_F^2,
\end{equation}
% where $\mathbf{P}_{x}, \mathbf{P}_{y}, \mathbf{P}_{z} \in\mathbb{R}^{kM}$ are the vectorized $x$-, $y$- and $z$-coordinates of the points in patches.
Let $\mathbf{P} = [\mathbf{P}_{x}, \mathbf{P}_{y}, \mathbf{P}_{z}] \in \mathbb{R}^{kM \times 3}$, and
$\sum_{\Theta \in \{x,y,z\}} \mathbf{P}_{\Theta}^{\top} \mathbf{L}_p \mathbf{P}_{\Theta}$ can be combined as $\textrm{tr}(\mathbf{P}^\top \mathbf{L}_p \mathbf{P})$.
$\mathbf{P}$ is related to denoised 3D samples $\mathbf{U} \in \mathbb{R}^{N \times 3}$ as follows:
\begin{equation}
\mathbf{P} = \mathbf{S}\mathbf{U}-\mathbf{C},
\end{equation}
where $\mathbf{S} \in \{0, 1\}^{kM \times N}$ is a sampling matrix to select points from point cloud $\mathcal{V}$ to form $M$ patches of $k$ 3D points each, and $\mathbf{C} \in \mathbb{R}^{kM \times 3}$ is for patch centering. 
Hence, the objective function can be rewritten as:
\begin{equation}
\underset{\mathbf{U}}{\text{min}} \quad \textrm{tr}((\mathbf{S}\mathbf{U}-\mathbf{C})^\top \mathbf{L}_p (\mathbf{S}\mathbf{U}-\mathbf{C}))+\mu \left\| \mathbf{V} - \mathbf{U} \right\|_F^2.
\label{eq:obj1}
\end{equation}
Now the questions that remain are: i) how to find local correspondence between connected patch pairs for graph construction, and ii) how to implement the numerical optimization. They are addressed in the next section.
% Next section will discuss the graph construction based on a patch distance measure for unstructured point cloud, and the algorithm implementation with spectral-domain interpretation and stability analysis.}
% Move to next section: Then, we propose a patch distance measure for unstructured point cloud that is efficiently computed and robust to noise.

\section{Algorithm Development}
\label{sec:algo}
In this section, we first propose a patch distance measure for graph construction, and then discuss the algorithm implementation.
Then we show that, with GLR, the algorithm is guaranteed with numerical stability and can be solved efficiently.

\subsection{Patch Distance Measure}

%A pair of same-size image patches on a regular 2D pixel grid naturally have one-to-one matching between pixels in the two patches, thus it is easy to compute image patch distance. 
%However, surface patches in 3D point cloud are unstructured, and we need to first find pairwise correspondences between points in the two patches and then compute the patch distance.

% \subsubsection{From Global Coordinate Ordering to Local Correspondence}

%Given the patch distance measure, the patch graph is constructed. To convert back to the point domain and connect the points,
%\blue{some content in section \ref{sec:laplacian} is moved here}

\subsubsection{Distance Measure in Continuous Domain}

\begin{figure}
\centering
\subfigure[]{\includegraphics[width=0.45\linewidth]{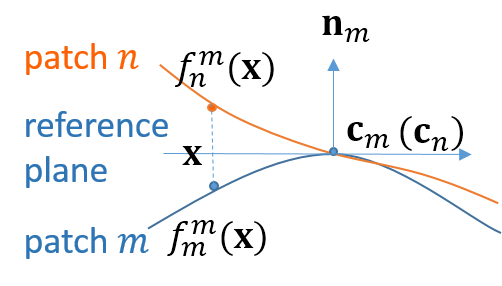}}
\subfigure[]{\includegraphics[width=0.45\linewidth]{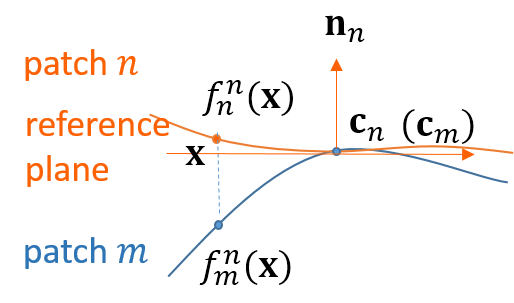}}
\caption{Distance measure with continuous surfaces with reference plane perpendicular to :(a) surface normal $\mathbf{n}_m$ at center of patch $m$, and (b) surface normal $\mathbf{n}_n$ at center of patch $n$.}
\label{fig:cont}
\end{figure} 

To measure the distance between the $m$-th patch and $n$-th patch, \textit{ideally} the two patches can be interpolated to two continuous surfaces, and the distance is calculated as the integral of the surface distance over a local domain around the patch center. 

To define the underlying surface, we first define a \textit{reference plane}. In Fig.\;\ref{fig:cont}(a), we examine a 2D case for illustration. 
The reference plane is tangent to the center $\mathbf{c}_m$ of patch $m$ (origin point) and perpendicular to the surface normal $\mathbf{n}_m$ at $\mathbf{c}_m$. 
Then the \textit{surface distance} for patch $m$ with respect to normal $\mathbf{n}_m$ is defined as a function $f_m^m(\mathbf{x})$, where $\mathbf{x}$ is a point on the reference plane, superscript $m$ indicates that the reference plane is perpendicular to $\mathbf{n}_m$, while the subscript $m$ indicates the function defines patch $m$. $f_m^m(\mathbf{x})$ is then the perpendicular distance from $\mathbf{x}$ to surface $m$ with respect to normal $\mathbf{n}_m$. Surface $n$ is similarly defined as $f_n^m(\mathbf{x})$.
Note that because the patches are centered, $\mathbf{c}_m = \mathbf{c}_n$ which is the origin, but their surface normals $\mathbf{n}_m$ and $\mathbf{n}_n$ are typically different.

The patch distance is then computed as  
\begin{equation}
d_{\overrightarrow{mn}} = \sqrt{\frac{1}{|\Omega_m|}\int_{\mathbf{x} \in \Omega_m} (f_m^m(\mathbf{x}) - f_n^m(\mathbf{x}))^2 d\mathbf{x}},
\end{equation}
where $\Omega_m$ is the local neighborhood at $\mathbf{c}_m$. $|\Omega_m|$ is the area of $\Omega_m$. $d_{\overrightarrow{mn}}$ denotes the distance measured with reference plane perpendicular to $\mathbf{n}_m$.

Note that different reference planes lead to different distance values, so we alternately use $\mathbf{n}_m$ and $\mathbf{n}_n$ to define the reference plane. Fig.\,\ref{fig:cont}(b) illustrates the computation of $d_{\overrightarrow{nm}}$ with reference plane perpendicular to $\mathbf{n}_n$.
\begin{equation}
d_{\overrightarrow{nm}} = \sqrt{\frac{1}{|\Omega_n|}\int_{\mathbf{x} \in \Omega_n} (f^n_m(\mathbf{x}) - f^n_n(\mathbf{x}))^2 d\mathbf{x}},
\end{equation}
where functions $f^n_m$ and $f^n_n$ define surfaces $m$ and $n$, respectively, and $\Omega_n$ the local neighborhood at $\mathbf{c}_n$.

$d_{mn}$ is then given as,
\begin{equation} \label{eq:dmn_2}
d_{mn} = \sqrt{\frac{d_{\overrightarrow{mn}}^2 + d_{\overrightarrow{nm}}^2}{2}}.
\end{equation}

\subsubsection{Distance Measure with Discrete Point Observation}

\begin{figure}
\centering
\subfigure[]{\includegraphics[width=0.48\linewidth]{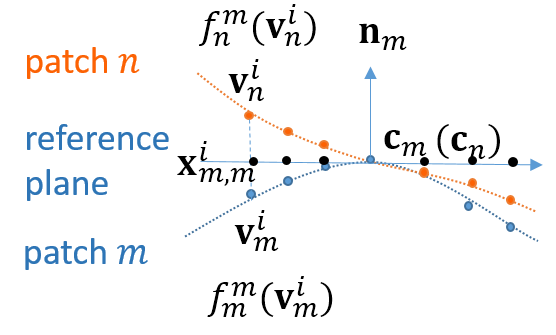}}
\subfigure[]{\includegraphics[width=0.48\linewidth]{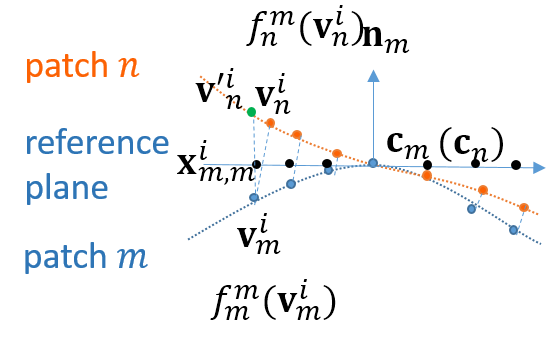}}
\caption{Distance measure with discrete patches with reference plane perpendicular to surface normal $\mathbf{n}_m$ at center of patch $m$. (a) Ideal case where the points with the same projection is connected. (b) A point in patch $m$ is connected with the closest point in patch $n$ in terms of projection distance.}
\label{fig:dist}
\end{figure} 

Since we only have discrete observations of the points on the patches, we instead measure the sum of the distances between points with the same projection on the reference plane. 

First, we compute $d_{\overrightarrow{mn}}$, where reference plane is perpendicular to the surface normal $\mathbf{n}_m$ at $\mathbf{c}_m$. 
%\red{gene: how do u compute surface normals for a patch of discrete points?}
Specifically, patch $m$ is composed of points $\{\mathbf{v}_m^i\}_{i=1}^k$, while patch $n$ is composed of $\{\mathbf{v}_n^i\}_{i=1}^k$.
%\blue{explanation for normal estimation}
The surface normal $\mathbf{n}_m$ is given by,
\begin{equation}
\underset{\mathbf{n}_m}{\text{min}} \quad \sum_{i=1}^k((\mathbf{v}_m^i)^{\top}\mathbf{n}_m)^2.
\end{equation}
It can be shown via Principal Component Analysis \cite{horn12} that the solution is the normalized eigenvector according to the smallest eigenvalue of the covariance matrix $\mathbf{Q}$ given by,
\begin{equation}
\mathbf{Q} = \frac{1}{k}\sum_{i=1}^k \mathbf{v}_m^i(\mathbf{v}_m^i)^{\top}.
\end{equation}
The same normal estimation method is used in PCL Library \cite{rusu20113d}.
We then project both $\{\mathbf{v}_m^i\}_{i=1}^k$ and $\{\mathbf{v}_n^i\}_{i=1}^k$ to the reference plane, and the projections are $\{\mathbf{x}_{m,m}^i\}_{i=1}^k$ and $\{\mathbf{x}_{n,m}^i\}_{i=1}^k$ respectively, where the second index in subscript indicates that the reference plane is perpendicular to $\mathbf{n}_m$.
The distances between $\mathbf{v}_m^i$ and $\mathbf{x}_{m,m}^i$ give the surface distance $f_m^m(\mathbf{v}_m^i)$, and the distances between $\mathbf{v}_n^i$ and $\mathbf{x}_{n,m}^i$ give $f_n^m(\mathbf{v}_n^i)$.
%$\mathbf{x}_m = \mathbf{v}_m - f_m(\mathbf{x}_m)\mathbf{n}_m$.

Ideally, for any $\mathbf{v}_m^i$ in patch $m$, there exists a point $\mathbf{v}_n^i$ in patch $n$ whose projection $\mathbf{x}_{n,m}^i = \mathbf{x}_{m,m}^i$ as shown in Fig.\;\ref{fig:dist}(a). 
However, in real dataset, $\mathbf{v}_m^i$ usually does not have a match in patch $n$ with exactly the same projection, as illustrated in Fig.\;\ref{fig:dist}(b). 
In this case, we replace the displacement value of $\mathbf{v'}_n^i$ (the green point in Fig.\,\ref{fig:dist}(b)), which has the same projection $\mathbf{x}_{m,m}^i$ as $\mathbf{v}_m^i$, with the value of its nearest neighbor $\mathbf{v}_n^i$ in patch $n$ in terms of the distance between their projections $\mathbf{x}_{m,m}^i$ and $\mathbf{x}_{n,m}^i$. 
%\red{gene: not sure what u r interpolating. u r using only ONE nearest neighbor? I was under the impression that u r interpolating using two nearest neighbors for a 1D case.}
Then $d_{\overrightarrow{mn}}$ is computed as,
\begin{equation}
d_{\overrightarrow{mn}} = \sqrt{\frac{1}{k}\sum_{i=1}^{k} (f^m_m(\mathbf{v}_m^i) - f^m_n(\mathbf{v}_n^i))^2}. 
\end{equation}

Similarly, to compute $d_{\overrightarrow{nm}}$, we define reference plane with $\mathbf{n}_n$, then compute the projections $\{\mathbf{x}_{m,n}^{i}\}_{i=1}^k$ and $\{\mathbf{x}_{n,n}^{i}\}_{i=1}^k$ and displacements $f^n_m(\mathbf{v}_m^i)$, $f^n_n(\mathbf{v}_n^i)$.
For each $\mathbf{v}_n^i$ in patch $n$, we match it to the closest point in patch $m$ in terms of projection distance.
Then $d_{\overrightarrow{nm}}$ is computed as,
\begin{equation}
d_{\overrightarrow{nm}} = \sqrt{\frac{1}{k}\sum_{i=1}^{k} (f^n_m(\mathbf{v}_m^i) - f^n_n(\mathbf{v}_n^i))^2}.
\end{equation}
The final distance is given as (\ref{eq:dmn_2}).
% \begin{equation}
% d_{mn} = \sqrt{\frac{d_{\overrightarrow{mn}}^2 + d_{\overrightarrow{nm}}^2}{2}}.
% \end{equation}

\subsubsection{Planar Interpolation}

\begin{figure}
\centering
        \subfigure[]{\includegraphics[width=0.30\linewidth]{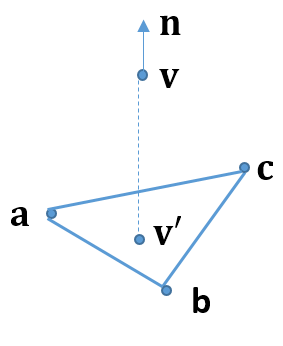}}
        \subfigure[]{\includegraphics[width=0.45\linewidth]{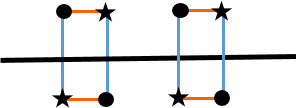}} \caption{(a) Interpolation for $\mathbf{v}$ on the plane $\mathbf{abc}$. (b) Patch connection based on projection (in blue) vs euclidean distance (in orange).}
\label{fig:project}
\end{figure} 

The pairwise correspondence is based on nearest neighbor replacement, though more accurate interpolation can be adopted. However, due to the large size of the point cloud, implementing interpolation for all the points can be expensive. 
Thus we use nearest-neighbor replacement when the distance between point pair is under a threshold $\tau$. 
%\red{gene: I am not sure what u mean by ``interpolation" here. typically interpolation means (weighted) averaging using multiple neighboring points. but u r using only one point, as far as I can understand. so instead of ``interpolation", u r doing nearest neighbor replacement? is that more appropriate?}
%\blue{yes it is in fact a replacement, revised accordingly.}
When the distance goes above $\tau$, we apply the interpolation method described as follows. 

As shown in Fig.\;\ref{fig:project}(a), for a point $\mathbf{v}$, to find its corresponding interpolation on the other patch, we find the three nearest points (also in terms of projection distance) to form a plane, and the interpolation $\mathbf{v}'$ is given by its projection along the normal vector $\mathbf{n}$ on the plane. It can be easily derived that the distance between $\mathbf{v}$ and $\mathbf{v}'$ is $\frac{\mathbf{n}^\top\mathbf{v} + d}{\mathbf{n}^\top \mathbf{n}_0}$ where $\mathbf{n}_0 = \vec{\mathbf{ab}} \times \vec{\mathbf{ac}}$ is the normal vector for the plane $\mathbf{abc}$, and $d = -\mathbf{n}_0^\top \mathbf{a}$. 

\subsubsection{Relation to Hausdorff Distance}
\blue{Hausdorff distance \cite{aspert2002mesh} is a widely used measure for comparing point clouds, which is derived from the Hausdorff distance for comparing the metric spaces of two manifolds and extended to deal with point clouds \cite{rockafellar2009variational}. The proposed patch distance measure is closely related to the modified Hausdorff distance (MHD) \cite{dubuisson1994modified}, which is a variant of Hausdorff distance. It decreases the impact of outliers and is more suitable for pattern recognition tasks. Specifically, MHD from the $m$-th patch and $n$-th patch is given as:
\begin{equation}
\mathrm{MHD}_{\overrightarrow{mn}} = \frac{1}{k}\sum_{i=1}^k \|\mathbf{v}_m^i - \mathbf{v}_n^i\|,
\end{equation}
where $\| \cdot \|$ is the Euclidean distance, $\mathbf{v}_n^i$ is the nearest neighbor of $\mathbf{v}_m^i$ in patch $n$ in terms of point position.
The major difference between MHD and our patch distance measure is that, we choose to use projection on the reference plane (\textit{e.g.} $\mathbf{x}_m^i$ in Fig.\;\ref{fig:dist}(a)) to find the correspondence, while MHD uses the point position (\textit{e.g.} $\mathbf{v}_m^i$ in Fig.\,\ref{fig:dist}(a)). }

Due to the use of projection, the proposed measure is more robust to noise than MHD. 
For example in Fig.\;\ref{fig:project}(b), the underlying surfaces for two patches are both planar, where the circle points belong to one patch and the star points belong to the other. 
The correct connections are between points along the vertical lines (in blue). 
This is accomplished by using projection on the reference plane. 
On the other hand, if the connection is decided by point position, then the resulting connections are erroneous (in orange) and thus lead to inefficient denoising. 
Therefore point connection based on projection is closer to the ground truth and more robust to noise. 

\subsection{Graph Construction}
\label{sec:laplacian}

Based on the above patch distance measure strategy, the connection between $m$-th and $n$-th patch is implemented as follows. If no interpolation is involved, the points in the $m$-th patch are connected with the nearest points in the $n$-th patch in terms of their projections on the reference plane decided by surface normal of patch $m$. 
Also, the points in the $n$-th patch are connected with the nearest points in the $m$-th patch in terms of their projections on the reference plane decided by surface normal of patch $n$. 
The edges are undirected and assigned the same weight $w_{mn}$ decided by $d_{mn}$ in (\ref{eq:w}). 

If interpolation is involved, for example in Fig.\,\ref{fig:project}(a), the weight $w_{va}$ between $\mathbf{v}$ and $\mathbf{a}$ is given by,
\begin{equation}
w_{va} = \frac{w_{mn} d_{va}}{d_{va} + d_{vb} + d_{vc}},
\end{equation}
where $w_{mn}$ is the weight between patch $m$ and $n$. Point $\mathbf{v}$ lies on patch $m$ and points $\mathbf{a}$, $\mathbf{b}$, $\mathbf{c}$ lie on patch $n$. $d_{va}$ is distance between $\mathbf{v}$ and $\mathbf{a}$, and similarly for $d_{vb}$ and $d_{vc}$.
To simplify the implementation, we limit the search range to be patches centered at the $K$-nearest patch centers, and evaluate patch distance between these $K$-nearest patches instead of all the patches in the point cloud.

In this way, the local correspondence is generated and the point domain graph is constructed, giving the graph Laplacian $\mathbf{L}_p$ in (\ref{eq:obj1}).

\subsection{Denoising Algorithm}
The optimization in (\ref{eq:obj1}) is non-convex because of $\mathbf{L}_p$'s dependency on patches in $\mathbf{P}$.  
To solve (\ref{eq:obj1}) approximately, we take an alternating approach, where in each iteration, we fix $\mathbf{L}_p$ and solve for $\mathbf{U}$, then update $\mathbf{L}_p$ given $\mathbf{U}$, and repeat until convergence. 

In each iteration, graph Laplacian $\mathbf{L}_p$ is easy to update using the previously discussed graph construction strategy. To optimize $\mathbf{U}$ for fixed $\mathbf{L}_p$, each of the $(x,y,z)$ coordinate is given by,
\begin{equation} \label{eq:lss}
(\mathbf{S}^\top \mathbf{L}_p \mathbf{S} + \mu \mathbf{I}) \mathbf{U}_{\Theta} = \mu \mathbf{V}_{\Theta} + \mathbf{S}^\top \mathbf{L}_p \mathbf{C}_{\Theta},
\end{equation}
where $\Theta \in \{x,y,z\}$ is the index for $(x,y,z)$ coordinates, and $\mathbf{I}$ is the identity matrix of the same size as $\mathbf{L}_p$. We iteratively solve the optimization until the result converges. The proposed algorithm is referred to as \textit{Graph Laplacian Regularized point cloud denoising} (GLR).
The algorithm is summarized in Algorithm \ref{algo:GLR}.

\begin{algorithm}[t]
\caption{Graph Laplacian Regularized Point Cloud Denoising}
\label{algo:GLR}
\begin{small}
\begin{algorithmic}[1]
\Require Noisy point cloud $\mathbf{V}$, patch center sampling rate $s$\%, patch size $k$, threshold $\tau$, max iteration number $r$ 
\Ensure Denoised point cloud $\mathbf{U}$ 
\State Initialize $\mathbf{U}^0 \gets \mathbf{V}$
\For{$i$ = 1 to $r$}  
    \State Sample $s$\% points from $\mathbf{U}^i$ as patch centers
    \State Find $k$ nearest neighbors of each patch center to form surface patches 
    \State Connect each patch center with $K$ nearest neighboring patch centers to give $\mathcal{E}$ 
    \For{$(m, n) \in \mathcal{E}$}
        \State Connect corresponding points between $m$-th and $n$-th patches and compute $\mathbf{L}_{mn}$
    \EndFor
    \State$\mathbf{L}_p \gets \sum_{(m,n)\in \mathcal{E}} \mathbf{S}_{mn}^{\top} \mathbf{L}_{mn} \mathbf{S}_{mn}$
    \State $\mathbf{U}^i_{\Theta} \gets (\mathbf{S}^\top \mathbf{L}_p \mathbf{S} + \mu \mathbf{I})^{-1}(\mu \mathbf{U}^{i-1}_{\Theta} + \mathbf{S}^\top \mathbf{L}_p \mathbf{C}_{\Theta})$, $\Theta \in \{x,y,z\}$
    \State End if $\mathbf{U}^i$ converges
\EndFor
\end{algorithmic}
\end{small}
\end{algorithm}

\subsection{Graph Spectral Analysis}

To impart intuition and demonstrate stability of our computation, in each iteration we can compute the optimal $x$-, $y$- and $z$-coordinates in (\ref{eq:obj1}) separately, resulting in the system of linear equations in (\ref{eq:lss}).
% \begin{align}
% (\mathbf{S}^\top \mathbf{L}_p \mathbf{S} + \mu \mathbf{I}) \mathbf{u}_x^* = \mu \; \mathbf{v}_x + \mathbf{S}^\top \mathbf{L}_p \mathbf{c}_x,
% \label{eq:objVec}
% \end{align}
% where $\mathbf{u}_x^*$, $\mathbf{v}_x$ and $\mathbf{c}_x$ represent the $x$-coordinates (first column) of $\mathbf{U}^*$, $\mathbf{V}$ and $\mathbf{C}$, respectively.
In Section\;\ref{sec:ld}, we assume that union of all $M$ patches covers all points in the point cloud $\mathcal{V}$, hence we can safely assume that $k M > N$. 

Because $\mathbf{S}$ is a sampling matrix, we can define $\mathbf{L} = \mathbf{S}^\top \mathbf{L}_p \mathbf{S}$ as a $N \times N$ \textit{principal sub-matrix}\footnote{A principal sub-matrix $B$ of an original larger matrix $A$ is one where the $i$-th row and column of $A$ are removed iteratively for different $i$.} of $\mathbf{L}_p$. 
Denote by $\lambda_1^{\mu} \leq \ldots \leq \lambda_N^{\mu}$ the eigenvalues of matrix $\mathbf{L} + \mu \mathbf{I}$.
The solution to (\ref{eq:lss}) can thus be written as:
\begin{align}
\mathbf{U}_{\Theta}^* = \boldsymbol{\Phi} \boldsymbol{\Sigma}^{-1} \boldsymbol{\Phi}^T
\left(  \mu \; \mathbf{V}_{\Theta} + \mathbf{S}^\top \mathbf{L}_p \mathbf{C}_{\Theta} \right),
\label{eq:solVec}
\end{align}
where $\boldsymbol{\Phi} \boldsymbol{\Sigma} \boldsymbol{\Phi}^T$ is an eigen-decomposition\footnote{Eigen-decomposition is possible because the target matrix $\mathbf{L} + \mu \mathbf{I}$ is real and symmetric.} of matrix $\mathbf{L} + \mu \mathbf{I}$; \textit{i.e.}, $\boldsymbol{\Phi}$ contains as columns eigenvectors $\boldsymbol{\phi}_1, \ldots, \boldsymbol{\phi}_N$, and $\boldsymbol{\Sigma}$ is a diagonal matrix containing eigenvalues on its diagonal. 
In \textit{graph signal processing} (GSP) \cite{shuman13}, eigenvalues and eigenvectors of a variational operator---$\mathbf{L} + \mu \mathbf{I}$ in our case---are commonly interpreted as graph frequencies and frequency components. 
$\boldsymbol{\Phi}^\top$ is thus an operator (called \textit{graph Fourier \blue{basis}})
% (called \textit{graph Fourier transform} (GFT)) 
that maps a graph-signal $\mathbf{x}$ to its GFT coefficients $\boldsymbol{\zeta} = \boldsymbol{\Phi}^\top \mathbf{x}$. 

Observing that $\boldsymbol{\Sigma}^{-1}$ in (\ref{eq:solVec}) is a diagonal matrix:
\begin{align}
\boldsymbol{\Sigma}^{-1} = \mathrm{diag} \left( 1/(\lambda_1^{\mu} + \mu), \ldots, 1/(\lambda_N^{\mu} + \mu) \right),
\label{eq:sigma}
\end{align}
we can thus interpret the solution $\mathbf{U}_{\Theta}^*$ in (\ref{eq:solVec}) as follows.
The noisy observation $\mathbf{V}_{\Theta}$ (offset by centering vector $\mathbf{C}_{\Theta}$) is transformed to the GFT domain via $\boldsymbol{\Phi}^\top$ and \textit{low-pass filtered} per coefficient according to (\ref{eq:sigma})---low-pass because weights $1/(\lambda_i^{\mu} + \mu)$ for low frequencies are larger than large frequencies $1/(\lambda_{j}^{\mu} + \mu)$, for $i < j$.
The fact that we are performing 3D point cloud denoising via graph spectral low-pass filtering should not be surprising.

\subsection{Numerical Stability via Eigen-Analysis}

We can also estimate the stability of the system of linear equations in (\ref{eq:solVec}) via the following eigen-analysis. 
During graph construction, an edge weight $w_{i,j}$ is computed using (\ref{eq:w}), which is upper-bounded by $1$. 
Denote by $\rho_{\max}$ the maximum degree of a node in the graph, which in general $\rho_{\max} \ll N$. 
According to the Gershgorin circle theorem \cite{horn12}, given a matrix $\mathbf{A}$, a Gershgorin disc $i$ has radius $r_i =  \sum_{j|j \neq i} |A_{i,j}|$ and center at $A_{i,i}$.
For a combinatorial graph Laplacian $\mathbf{L}_p$, the maximum Gershgorin disc radius is the maximum node degree multiplied by the maximum edge weight, which is $\rho_{\max}$. 
Further, the diagonal entry $L_{i,i} = - \sum_{j|j \neq i} L_{i,j}$ for positive edge weights, which equals $r_i$. 
Thus all Gershgorin discs for a combinatorial graph Laplacian matrix have left-ends located at $0$. 
By the Gershgorin circle theorem, all eigenvalues have to locate inside the union of all Gershgorin discs.
This means that the maximum eigenvalue $\lambda^p_{\max}$ for $\mathbf{L}_p$ is upper-bounded by twice the radius of the largest possible disc, which is $2 \rho_{\max}$.

Now consider principal sub-matrix $\mathbf{L}$ of original matrix $\mathbf{L}_p$. 
By the eigenvalue interlacing theorem, large eigenvalue $\lambda_{\max}$ for $\mathbf{L}$ is upper-bounded by $\lambda^p_{\max}$ of $\mathbf{L}_p$. 
For matrix $\mathbf{L} + \mu \mathbf{I}$, the smallest eigenvalue $\lambda^{\mu}_{\min} \geq \mu$, because: i) $\mu \mathbf{I}$ shifts all eigenvalues of $\mathbf{L}$ to the right by $\mu$, and ii) $\mathbf{L}$ is PSD due to eigenvalue interlacing theorem and the fact that $\mathbf{L}_p$ is PSD. 
We can thus conclude that the condition number\footnote{Assuming $l_2$-norm is used and the matrix is normal, then the condition number is defined as the ratio $\lambda_{\max}/\lambda_{\min}$.} $C$ of matrix $\mathbf{L} + \mu \mathbf{I}$ on the left-hand side of (\ref{eq:lss}) can be upper-bounded as follows:
\begin{align}
C \leq \frac{2 \rho_{\max} + \mu}{\mu}.
\end{align}
Hence for sufficiently small $\rho_{\max}$, the linear system of equations in (\ref{eq:lss}) has a stable solution, and can be efficiently solved using indirect methods like preconditioned conjugate gradient (PCG).

\subsection{Complexity Analysis}

The complexity of the algorithm depends on two main procedures: one is the patch-based graph construction, and the other is in solving the system of linear equations.

For graph construction, for the $M$ patches, the $K$-nearest patches to be connected can be found in $O(KM\log M)$ time. Then for $k$-point patch distance measure, each pair takes $O(k\log k)$; with $MK$ pairs, the complexity is $O(kMK\log k)$ in total. For the system of linear equations, it can be solved efficiently with PCG based methods, with complexity of $O(kMK\sqrt{C})$ \cite{shewchuk1994introduction}. 
Finally, if GLR runs for a maximum of $r$ iterations, the total time complexity will be $ O(r(KM\log M + kMK\log k + kMK\sqrt{C})) \approx O(rKM(\log M + k\log k + k\sqrt{C}))$.

The parameters that can be adjusted for the complexity reduction are patch center sampling density, patch graph neighborhood size and patch size.
Details about the parameter setting and complexity comparison with other existing schemes are given in Section \ref{sec:result}.

\section{Experimental Results}
\label{sec:result}
\begin{figure*}[t]
\centering
       \subfigure[]{\includegraphics[width=0.30\linewidth]{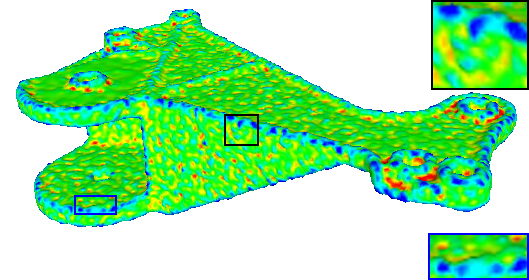}}
        \subfigure[]{\includegraphics[width=0.30\linewidth]{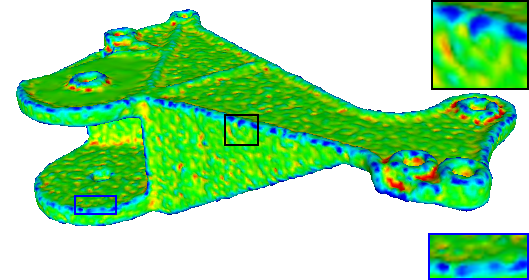}}
        \subfigure[]{\includegraphics[width=0.30\linewidth]{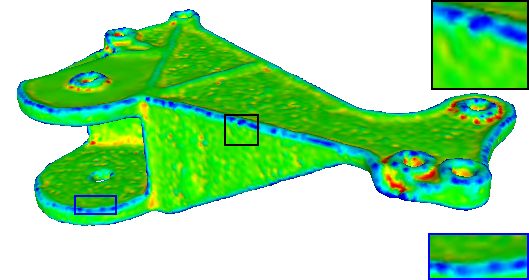}} 
\caption{\textit{Daratech} model ($\sigma=0.02$). Surface reconstruction with (a) noisy input, and denoising results of the proposed GLR after (b) iteration 1 and (c) iteration 3 (final output), colorized by mean curvature.}
\label{fig:daratech}
\end{figure*}

\begin{figure*}[t]
\centering 
        \subfigure[]{\includegraphics[width=0.30\linewidth]{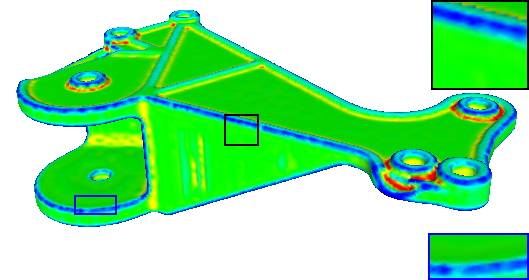}}
        \subfigure[]{\includegraphics[width=0.30\linewidth]{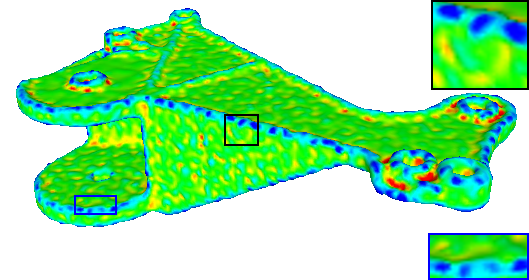}} 
        \subfigure[]{\includegraphics[width=0.30\linewidth]{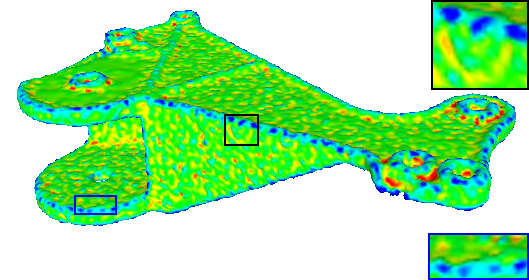}} 
        \\
        \subfigure[]{\includegraphics[width=0.30\linewidth]{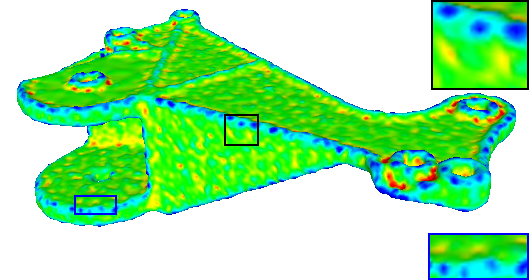}} 
        \subfigure[]{\includegraphics[width=0.30\linewidth]{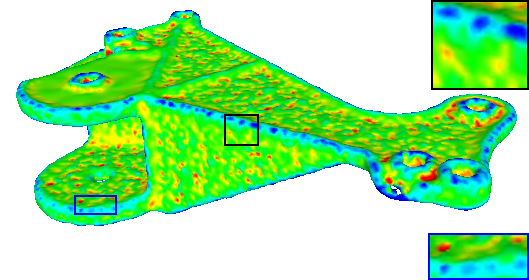}}
        \subfigure[]{\includegraphics[width=0.30\linewidth]{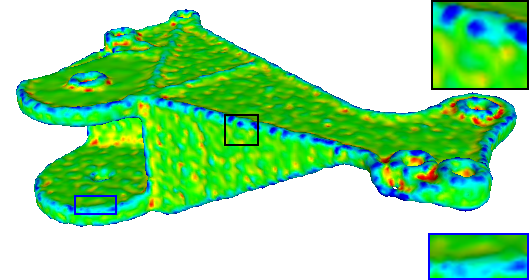}}
\caption{\textit{Daratech} model ($\sigma=0.02$). Surface reconstruction with (a) ground truth, and denoising results of (b) APSS, (c) AWLOP, (d) NLD, (e) MRPCA and (f) LR, colorized with mean curvature.}
\label{fig:daratech_comp}
\end{figure*}

The proposed scheme GLR is compared with existing works:  APSS \cite{guennebaud2007algebraic}, RIMLS \cite{oztireli2009feature}, AWLOP \cite{huang2013edge}, non-local denoising (NLD) algorithm \cite{deschaud2010point} and the state-of-the-art MRPCA \cite{matteipoint} \blue{and LR \cite{sarkar2018structured}}.
APSS and RIMLS are implemented with MeshLab software \cite{cignoni2008meshlab}, AWLOP is implemented with EAR software \cite{huang2013edge}, MRPCA source code is provided by the author, NLD \blue{and LR} are implemented by ourselves in MATLAB. 
\blue{We first empirically tune parameters on a small dataset with 8 models, then generalize the parameter setting learned from the small dataset to a larger dataset, \textit{i.e.}, 100 models from the ShapeNetCore dataset \cite{shapenet2015} for validation. Comparison with existing methods on both dataset are detailed as follows.}

\subsection{Evaluation Metrics}

Before the discussion of experimental performance, we first introduce the three evaluation metrics for point cloud denoising. 
Suppose the ground-truth and predicted point clouds are $\mathcal{U} = \{\mathbf{u}_i\}_{i=1}^{N_1}$, $\mathcal{V} = \{\mathbf{v}_i\}_{i=1}^{N_2}$, where $\mathbf{u}_i, \mathbf{v}_i \in \mathbb{R}^3$. The point clouds can be of different sizes, \textit{i.e.}, $N_1$ and $N_2$ may be unequal. The metrics are defined as follows.
\begin{enumerate}
\item mean-square-error (MSE): We first measure the average of the squared Euclidean distances between ground truth points and their closest denoised points, and also between the denoised points and their closest ground truth points, then take the average between the two measures to compute MSE which is given as
\begin{align} \nonumber
\mathrm{MSE} \: = \: &\frac{1}{2N_1}\sum\limits_{\mathbf{u}_i \in \mathcal{U}} \min_{\mathbf{v}_j \in \mathcal{V}} \|\mathbf{u}_i-\mathbf{v}_j\|_2^2 \\ 
&+ \frac{1}{2N_2}\sum\limits_{\mathbf{v}_i \in \mathcal{V}} \min_{\mathbf{u}_j \in \mathcal{U}} \|\mathbf{v}_i-\mathbf{u}_j\|_2^2
\end{align}

\item signal-to-noise ratio (SNR): SNR is measured in dB given as 
\begin{equation}
\mathrm{SNR} = 10 \log \frac{1/N_2\sum_{\mathbf{v}_i \in \mathcal{V}} \|\mathbf{v}_i\|_2^2 }{\mathrm{MSE}}
\end{equation}
\item mean city-block distance (MCD): MCD is similar to MSE with $l_2$ norm replaced with $l_1$ norm, given as
\begin{align} \nonumber
\mathrm{MCD} \: = \: &\frac{1}{2N_1}\sum\limits_{\mathbf{u}_i \in \mathcal{U}} \min_{\mathbf{v}_j \in \mathcal{V}} |\mathbf{u}_i-\mathbf{v}_j| \\ 
&+ \frac{1}{2N_2}\sum\limits_{\mathbf{v}_i \in \mathcal{V}} \min_{\mathbf{u}_j \in \mathcal{U}} |\mathbf{v}_i-\mathbf{u}_j|
\end{align}

\end{enumerate}

\subsection{Parameter Tuning}
\label{sec:parameter}
8 models are used for parameter tuning, including \textit{Anchor}, \textit{Bimda}, \textit{Bunny}, \textit{Daratech}, \textit{DC}, \textit{Fandisk}, \textit{Gargoyle} and \textit{Lordquas} provided in \cite{rosman2013patch} and \cite{matteipoint}. The models are around 50000 in size. 
\blue{Gaussian noise with zero-mean is added to the 3D positions of each point cloud, where the standard deviation is set proportional to the signal scale as commonly used in point cloud denoising works \cite{matteipoint,rosman2013patch}.
We first compute the diameter of the point cloud, which is the maximum distance among 200 points sampled from the point cloud using farthest point sampling \cite{eldar1997farthest}. Then the standard deviation of the additive Gaussian noise is the multiplication of the diameter and $\sigma$, where $\sigma =$ 0.02, 0.03, 0.04.}

\subsubsection{\blue{Parameters in Optimization Formulation}}
\label{sec:optpara}
\blue{For implementation of the proposed GLR, we need to tune the parameter $\mu$ for balancing the data fidelity term and the regularization term in (\ref{eq:mu}), and $\epsilon$ for weighting the edge between connected patches in (\ref{eq:kernelfunc}).}

\blue{In (\ref{eq:mu}), the GLR regularization reflects the prior expectation of signal smoothness on the graph \cite{shuman13} which can be estimated from the dataset for parameter tuning.
Meanwhile, the data term measures the noise variance, and its ratio to expected signal smoothness is found to be similar for different models given the same noise level $\sigma$. This is because the noise standard variance is set proportional to signal scale as explained above. Therefore, given noise level $\sigma$, $\mu$ is tuned on the 8 models and generalizes to other models. 
% Therefore, $\mu$ is not tuned for each model separately, but for the 8 models jointly.
Specifically, $\mu = 25(\exp(\mathrm{iteration}/r)-1)$ which increases along the iterations, where $r=4,7,12$ for $\sigma$ = 0.02, 0.03, 0.04, respectively. }

\blue{From (\ref{eq:kernelfunc}), we can see $\epsilon$ should be proportional to square root of standard deviation of the patch distances, \textit{i.e.}, $\xi = \sqrt{\mathrm{std}(d^2)}$ with a moderate scale, where $\mathrm{std}$ is the standard deviation, $d$ denotes the distance of patch pair.
% If $\epsilon$ is set too large, the edge weight is insensitive to the patch distance, thus the filtering is not edge-aware; 
% if set too small, the edge weight is so sensitive to distance that only very close neighbors get filtered, taking more iterations to reach convergence in result.
Through testing on the 8 models, we empirically set $\epsilon = 0.5\xi$.}
% $\epsilon$ is set to be $0.07$ for iteration 1, $0.05$ for iteration 2, and $0.03$ for the iterations afterwards.
% Parameters that require tuning for better performance include $\mu$ in the objective and $\epsilon$ in deciding the graph edge weight, which will be discussed in Section \ref{sec:iters} and \ref{sec:numerical} in detail for particular models. 

\subsubsection{\blue{Parameters for Performance and Speed Balance}}
\label{sec:optpara2}
To speed up the implementation, we take 50\% of the points as the patch centers, with the farthest point sampling \cite{eldar1997farthest} to assure spatially uniform selection.
The planar interpolation threshold $\tau$ is set to 1, which is large enough to ensure most points are connected using nearest-neighbor replacement for efficient implementation. The maximum iteration number $r$ is set to 15.
\blue{To find the proper value for the search window size $K$ and the patch size $k$, we study the performance sensitivity to $K$ and $k$. We set $K = 4,8,12,16,20$ and $k = 15, 30, 60, 120$, and test on the 8 models.
The average MSE results are shown in Table \ref{tab:wink} and \ref{tab:pk}. }

\blue{With larger search range $K$, each patch is more likely to find similar patches to get connected, so the results get better though the performance converges when $K$ reaches 16.
On the other hand, the runtime increases as $K$ gets larger, so we choose $K$ to be 16 to balance the performance and runtime.
With small $k$, the patch size is too small to capture salient features, so the filtering cannot distinguish patch similarity and is not edge-aware; with large $k$, the patch contains too many salient features and the dimension of the patch manifold increases, thus the low-dimensional manifold model assumption is invalid.
Therefore we choose $k=30$ as a suitable patch size which provides the best results in Table \ref{tab:pk}.}
% The search window size $K$ is fixed to be 16, and the patch size $k$ is fixed to be 30. 

\begin{table}[htbp]
  \centering
  \caption{\blue{MSE Results and Runtime (sec) of Different Setting for $K$}}
    \begin{tabular}{|r||r|r|r|r|r|}
    \hline
    Noise Level & $K=4$   & $K=8$   & $K=12$  & $K=16$  & $K=20$  \\
    \hline
    \hline
    0.02  & 0.147 & 0.145 & 0.143 & \textbf{0.142} & 0.143  \\
    \hline
    0.03  & 0.173 & 0.171 & 0.166 & 0.165 & \textbf{0.164}  \\
    \hline
    0.04  & 0.196 & 0.190 & 0.184 & \textbf{0.181} & \textbf{0.181}  \\
    \hline
    runtime & 131.6 & 227.4 & 323.7 & 399.0 & 496.2  \\
    \hline
    \end{tabular}%
  \label{tab:wink}%
\end{table}%

\begin{table}[htbp]
  \centering
  \caption{\blue{MSE Results and Runtime (sec) of Different Setting for $k$}}    \begin{tabular}{|r||r|r|r|r|}
    \hline
    \multicolumn{1}{|l||}{Noise Level} & \multicolumn{1}{l|}{$k=15$} & \multicolumn{1}{l|}{$k=30$} & \multicolumn{1}{l|}{$k=60$} & \multicolumn{1}{l|}{$k=120$}   \\
    \hline
    \hline
    0.02  & 0.177 & \textbf{0.142} & 0.145 & 0.170   \\
    \hline
    0.03  & 0.206 & \textbf{0.165} & 0.170 & 0.285   \\
    \hline
    0.04  & 0.229 & \textbf{0.181} & 0.292 & 0.655   \\
    \hline
    runtime & 262.8 & 399.0 & 852.7 & 2535.2   \\
    \hline
    \end{tabular}%
  \label{tab:pk}%
\end{table}%

\subsubsection{Objective Comparison with Existing Methods}

MSE, SNR and MCD results comparison with different methods on the 8 models are shown in Table \ref{tab:modelMSE}, \ref{tab:modelSNR}, and \ref{tab:modell1}, where the numbers showing the best performance are highlighted in bold. 

\blue{For parameter settings of competing methods, NLD and LR follow the default settings in the corresponding papers; the rest of the methods require manual parameter tuning, and optimal parameters vary for different models as shown in Table \ref{tab:otherpara}. Parameters not shown in Table \ref{tab:otherpara} follow the default setting in the software.}

GLR achieves the best results on average in all three metrics and all noise levels. 
In terms of MSE, GLR outperforms the second best scheme by 0.009, 0.008 and 0.009 for $\sigma=$ 0.02, 0.03, 0.04; for SNR, GLR outperforms the second best by 0.72 dB, 0.96 dB, 0.75 dB for $\sigma=$ 0.02, 0.03, 0.04; for MCD, GLR outperforms the second best by 0.013, 0.013, 0.011 for $\sigma=$ 0.02, 0.03, 0.04. 
APSS and MRPCA are usually the second and the third best among different methods. APSS never achieves the best result for one single model, but on average outperforms others because the local sphere fitting provides stable results. 
For MRPCA, it sometimes outperforms GLR but on average is only ranked third because of the unstable performance since the sparsity regularization is likely to generate extra features \cite{matteipoint}. 
In contrast, the proposed GLR not only has stable performance due to the robustness to high noise level, but also outperforms the other schemes overall, validating the effectiveness of LDMM.
\blue{The patch-similarity based LR is not among the top methods because the patch extraction procedure causes fine detail lose as discussed in Section \ref{sec:related}, but outperforms the non-local means based NLD, validating the effectiveness of using patch self-similarity.}

\begin{table*}[htbp]
  \centering \scriptsize
  \caption{\blue{Parameter Setting of Competing Methods for Different Models and Noise Levels}}
    \begin{tabular}{|l|l|l|l|l|l|l|l|l|l|}
    \hline
    \multirow{2}[1]{*}{Methods} & \multicolumn{1}{l|}{\multirow{2}[1]{*}{Parameters}} & \multicolumn{8}{c|}{Parameter Setting for Different $\sigma$ 0.02\,$|$\,0.03\,$|$\,0.04}  \\
\cline{3-10}          &       & Anchor         & Bimba            & Bunny          & Daratech       & DC             & Fandisk         & Gargoyle       & Lordquas  \\
    \hline
    APSS  & \multicolumn{1}{l|}{filter scale} & \multicolumn{1}{l|}{5\,$|$\,5\,$|$\,5} & \multicolumn{1}{l|}{5\,$|$\,10\,$|$\,10} & \multicolumn{1}{l|}{5\,$|$\,5\,$|$\,5} & \multicolumn{1}{l|}{3\,$|$\,3\,$|$\,3} & \multicolumn{1}{l|}{5\,$|$\,5\,$|$\,5} & \multicolumn{1}{l|}{4\,$|$\,5\,$|$\,8} & \multicolumn{1}{l|}{4\,$|$\,4\,$|$\,4} & \multicolumn{1}{l|}{6\,$|$\,6\,$|$\,6}  \\
    \hline
    RIMLS & \multicolumn{1}{l|}{filter scale} & \multicolumn{1}{l|}{7\,$|$\,7\,$|$\,7} & \multicolumn{1}{l|}{5\,$|$\,12\,$|$\,12} & \multicolumn{1}{l|}{5\,$|$\,5\,$|$\,5} & \multicolumn{1}{l|}{3\,$|$\,3\,$|$\,3} & \multicolumn{1}{l|}{5\,$|$\,5\,$|$\,5} & \multicolumn{1}{l|}{5\,$|$\,8\,$|$\,8} & \multicolumn{1}{l|}{5\,$|$\,5\,$|$\,5} & \multicolumn{1}{l|}{6\,$|$\,6\,$|$\,6}  \\
    \hline
    \multirow{2}[1]{*}{AWLOP} & \multicolumn{1}{l|}{repulsion force} & \multicolumn{1}{l|}{0.3\,$|$\,0.3\,$|$\,0.3} & \multicolumn{1}{l|}{0.3\,$|$\,0.5\,$|$\,0.5} & \multicolumn{1}{l|}{0.3\,$|$\,0.3\,$|$\,0.3} & \multicolumn{1}{l|}{0.3\,$|$\,0.3\,$|$\,0.3} & \multicolumn{1}{l|}{0.3\,$|$\,0.3\,$|$\,0.3} & \multicolumn{1}{l|}{0.3\,$|$\,0.3\,$|$\,0.3} & \multicolumn{1}{l|}{0.3\,$|$\,0.3\,$|$\,0.3} & \multicolumn{1}{l|}{0.3\,$|$\,0.3\,$|$\,0.5}  \\
    \cline{2-10}
          & \multicolumn{1}{l|}{iteration} & \multicolumn{1}{l|}{2\,$|$\,2\,$|$\,2} & \multicolumn{1}{l|}{2\,$|$\,10\,$|$\,10} & \multicolumn{1}{l|}{2\,$|$\,2\,$|$\,2} & \multicolumn{1}{l|}{2\,$|$\,2\,$|$\,2} & \multicolumn{1}{l|}{2\,$|$\,2\,$|$\,2} & \multicolumn{1}{l|}{2\,$|$\,2\,$|$\,2} & \multicolumn{1}{l|}{2\,$|$\,2\,$|$\,2} & \multicolumn{1}{l|}{2\,$|$\,2\,$|$\,10}  \\
    \hline
  \multirow{2}[1]{*}{MRPCA}  & \multicolumn{1}{l|}{data fitting} & \multicolumn{1}{l|}{1\,$|$\,1\,$|$\,1} & \multicolumn{1}{l|}{1\,$|$\,4\,$|$\,4} & \multicolumn{1}{l|}{1\,$|$\,1\,$|$\,1} & \multicolumn{1}{l|}{1\,$|$\,1\,$|$\,1} & \multicolumn{1}{l|}{0.01\,$|$\,0.01\,$|$\,0.01} & \multicolumn{1}{l|}{1\,$|$\,1\,$|$\,1} & \multicolumn{1}{l|}{1\,$|$\,1\,$|$\,1} & \multicolumn{1}{l|}{1\,$|$\,1\,$|$\,1}  \\
    \cline{2-10}
          & \multicolumn{1}{l|}{iteration} & \multicolumn{1}{l|}{6\,$|$\,6\,$|$\,6} & \multicolumn{1}{l|}{6\,$|$\,1\,$|$\,1} & \multicolumn{1}{l|}{6\,$|$\,6\,$|$\,6} & \multicolumn{1}{l|}{1\,$|$\,1\,$|$\,1} & \multicolumn{1}{l|}{2\,$|$\,2\,$|$\,2} & \multicolumn{1}{l|}{6\,$|$\,6\,$|$\,6} & \multicolumn{1}{l|}{2\,$|$\,2\,$|$\,2} & \multicolumn{1}{l|}{3\,$|$\,3\,$|$\,3}  \\
    \hline
    % NLD   &       &       &       &       &       &       &       &       &   \\
    % \hline
    % LR    &       &       &       &       &       &       &       &       &   \\
    % \hline
    \end{tabular}%
  \label{tab:otherpara}%
\end{table*}%

\begin{table*}[thb]
\centering
\caption{MSE Results of Different Methods on Small Dataset with Three Noise Levels}
\begin{tabular}{|c|c|c|c|c|c|c|c|c|c|c|} 
\hline
Noise level         & Methods & Anchor         & Bimba    & Bunny  & Daratech       & DC     & Fandisk         & Gargoyle       & Lordquas       & Average  \\ 
\hline\hline
\multirow{8}{*}{$\sigma$ = 0.02} & Noisy   & 0.259  & 0.0191   & 0.247  & 0.245  & 0.237  & 0.0258  & 0.257  & 0.224  &0.189  \\ 
\cline{2-11}
   & APSS    & 0.208  & 0.0131   & 0.198  & 0.203  & 0.186  & 0.0201  & 0.208  & 0.171  &0.151  \\ 
\cline{2-11}
   & RIMLS   & 0.212  & 0.0169   & 0.208  & 0.209  & 0.198  & 0.0196  & 0.217  & 0.183  & 0.158         \\ 
\cline{2-11}
   & AWLOP   & 0.237  & \textbf{0.0110}   & 0.223  & 0.228  & 0.211  & 0.0191  & 0.230  & 0.196  &0.169  \\ 
\cline{2-11}
   & NLD     & 0.231  & 0.0174   & 0.220  & 0.222  & 0.206  & 0.0208  & 0.230  & 0.190  &0.167  \\ 
\cline{2-11}
   & MRPCA   & 0.202  & 0.0154   & 0.213  & 0.225  & 0.189  & \textbf{0.0164} & 0.215  & 0.171  &  0.156        \\ 
\cline{2-11}        & LR   & 0.228 & 0.0133 & 0.220 & 0.213 & 0.206 & 0.0173 & 0.240 & 0.180 & 0.165         \\ 
% \cline{2-11}
%   & GLR     & \textbf{0.190} & \textbf{0.00968} & \textbf{0.187} & \textbf{0.201} & \textbf{0.179} & 0.0173  & \textbf{0.204} & \textbf{0.164} & \textbf{0.144}         \\
\cline{2-11}
   & GLR     & \textbf{0.189} & 0.0120 &\textbf{ 0.183} & \textbf{0.197} &\textbf{0.177} & 0.0173 & \textbf{0.202} & \textbf{0.162} & \textbf{0.142}       \\
\hline\hline
\multirow{8}{*}{$\sigma$ = 0.03} & Noisy   & 0.321  & 0.0257   & 0.309  & 0.304  & 0.292  & 0.0326  & 0.319  & 0.274  &  0.235        \\ 
\cline{2-11}
   & APSS    & 0.238  & 0.0196   & 0.228  & 0.242  & 0.210  & 0.0234  & 0.239  & 0.188  &  0.173        \\ 
\cline{2-11}
   & RIMLS   & 0.244  & 0.0213   & 0.241  & 0.255  & 0.225  & 0.0252  & 0.251  & 0.203  &  0.183        \\ 
\cline{2-11}
   & AWLOP   & 0.278  & \textbf{0.0133}  & 0.266  & 0.264  & 0.246  & 0.0218  & 0.270  & 0.226  & 0.198         \\ 
\cline{2-11}
   & NLD     & 0.265  & 0.0245   & 0.255  & 0.258  & 0.235  & 0.0285  & 0.262  & 0.217  & 0.193         \\ 
\cline{2-11}
   & MRPCA   & 0.230  & 0.0233   & 0.238  & 0.262  & 0.210  & 0.0239  & 0.241  & 0.187  &    0.177      \\
\cline{2-11}
   & LR   & 0.246 & 0.0209 & 0.237 & 0.252 & 0.221 & 0.0210 & 0.257 & 0.193 & 0.181       \\ 
% \cline{2-11}
%   & GLR     & \textbf{0.219} & 0.0138   & \textbf{0.216} & \textbf{0.238} & \textbf{0.202} & \textbf{0.0210} & \textbf{0.233} & \textbf{0.177} &    \textbf{0.165}      \\ 
\cline{2-11}
   & GLR     & \textbf{0.217} & 0.0147 & \textbf{0.217} & \textbf{0.238} & \textbf{0.203} & \textbf{0.0190} & \textbf{0.233} & \textbf{0.176} & \textbf{0.165}       \\
\hline\hline
\multirow{8}{*}{$\sigma$ = 0.04} & Noisy   & 0.372  & 0.0324   & 0.356  & 0.348  & 0.338  & 0.0391  & 0.368  & 0.318  &   0.271       \\ 
\cline{2-11}
   & APSS    & 0.254  & 0.0200   & 0.244  & 0.282  & 0.227  & 0.0289  & 0.262  & 0.201  &   0.190       \\ 
\cline{2-11}
   & RIMLS   & 0.263  & 0.0250   & 0.266  & 0.308  & 0.254  & 0.0314  & 0.277  & 0.219  &    0.205      \\ 
\cline{2-11}
   & AWLOP   & 0.306  & \textbf{0.0151}  & 0.291  & 0.286  & 0.270  & 0.0240 & 0.297  & 0.218  &0.213  \\ 
\cline{2-11}
   & NLD     & 0.297  & 0.0316   & 0.285  & 0.295  & 0.269  & 0.0372  & 0.294  & 0.252  &    0.220      \\ 
\cline{2-11}
   & MRPCA   & 0.242  & 0.0306   & 0.248  & 0.288  & \textbf{0.223}  & 0.0345  & \textbf{0.257}  & 0.199  &   0.190       \\
\cline{2-11}
   & LR   &0.259 & 0.0313 & 0.249 & 0.283 & 0.234 & 0.0297 & 0.269 & 0.204 & 0.195       \\ 
% \cline{2-11}
%   & GLR     & \textbf{0.241} & 0.0195   & \textbf{0.235} & \textbf{0.270} & \textbf{0.221} & 0.0241  & \textbf{0.254} & \textbf{0.186} &  \textbf{ 0.181}       \\
\cline{2-11}
   & GLR     & \textbf{0.228} & 0.0175 & \textbf{0.234} & \textbf{0.276} & 0.228 & \textbf{0.0229} & \textbf{0.257} & \textbf{0.187} &\textbf{ 0.181}       \\
\hline
\end{tabular}
\label{tab:modelMSE}
\end{table*}

\subsubsection{Visual Comparison with Existing Methods}
\label{sec:iters}
Here we demonstrate the results using the model \textit{Daratech} in Fig.\,\ref{fig:daratech_comp}(a) with $\sigma = 0.02$ shown in Fig.\,\ref{fig:daratech}(a). 
For better visualization, we demonstrate surfaces created from the point clouds with screened Poisson surface reconstruction algorithm \cite{kazhdan2013screened}, and colorize the points using the mean curvature calculated from APSS implemented in MeshLab software.

The surface reconstruction of GLR denoising results after 1\textit{st} and 3\textit{rd} iteration are shown in Fig.\,\ref{fig:daratech}(b) and (c), which demonstrate the iterative recovery of the point cloud.
The result converges fast and we do not show the result after iteration 3 since it already converges.

The surface patches in black and blue rectangles are enlarged and placed at the upper-right and lower-right corners to show structural details. The underlying plane (with curvature in green) and fold (with curvature in blue) are gradually recovered, smoothing out the noise on the plane while maintaining the edges. 

The comparison with other schemes is shown in Fig.\,\ref{fig:daratech_comp}. APSS in Fig.\,\ref{fig:daratech_comp}(b) generates relatively smoother surface than others as shown in the blue rectangles. However, the underlying true structure is not recovered due to the limitation of local operation, resulting in uneven planes and over-smoothed folds. RIMLS shows similar results as APSS thus is not shown in Fig.\,\ref{fig:daratech_comp}.

We observe that NLD in Fig.\,\ref{fig:daratech_comp}(d) has similar results as APSS and RIMLS, since the features used for similarity computation in NLD are based on the polynomial coefficients of the MLS surface. Moreover, NLD only takes one pass instead of multiple iterations since more iterations worsen its result as reported in \cite{deschaud2010point}, so the noise is not satisfactorily removed.
Though NLD and GLR both belong to the non-local category of methods, NLD is based on the non-local means scheme and is not collaboratively denoising the patches, thus also suffers from the drawback of local operation. 

AWLOP results in Fig.\,\ref{fig:daratech_comp}(c) have non-negligible noise. AWLOP is based on normal estimation, so the results indicate that AWLOP fails to estimate the normal at high noise level, and the noisy features may be regarded as sharp features and preserved.

MRPCA in Fig.\,\ref{fig:daratech_comp}(e) is not providing satisfying results, where the fold is already smoothed out but the plane is still uneven as shown in the blur rectangle.
LR in Fig.\,\ref{fig:daratech_comp}(f) generates relatively smoother planes than others shown in the blue rectangle but noise is still not fully removed.
For the proposed GLR in Fig.\,\ref{fig:daratech}(c), the result is visually better, preserving the plane and folding structures without over-smoothing.

% Illustrations of the \textit{Daratech} model from another perspective are shown in Fig.\,\ref{fig:daratech_comp2} for a more comprehensive evaluation. The proposed method exhibits sharper edges along the fold and smoother planes, while the others still have negligible noise.

% \begin{figure*}[t]
% \centering 
%         \subfigure[]{\includegraphics[width=0.24\linewidth]{eps/daratech_002_ny_v1_show.png}}
%         \subfigure[]{\includegraphics[width=0.24\linewidth]{eps/daratech_002_apss_v1_show.png}} 
%         \subfigure[]{\includegraphics[width=0.24\linewidth]{eps/daratech_002_rimls_v1_show.png}}
%         \subfigure[]{\includegraphics[width=0.24\linewidth]{eps/daratech_002_awlop_v1_show.png}} 
%         \\      
%         \subfigure[]{\includegraphics[width=0.24\linewidth]{eps/daratech_002_nld_v1_show.png}} 
%         \subfigure[]{\includegraphics[width=0.24\linewidth]{eps/daratech_002_mrpca_v1_show.png}} 
%         \subfigure[]{\includegraphics[width=0.24\linewidth]{eps/daratech_002_glr_v1_show.png}}
%         \subfigure[]{\includegraphics[width=0.24\linewidth]{eps/daratech_002_gt_v1_show.png}}
% \caption{\textit{Daratech} model ($\sigma=0.02$) from another view. Surface reconstruction with (a) noisy input, denoising results of (b) APSS, (c) RIMLS, (d) AWLOP, (e) NLD, (f) MRPCA, (g) proposed GLR, and (h) ground truth, colorized with mean curvature.}
% \label{fig:daratech_comp2}
% \end{figure*}

\begin{figure*}[t]
\centering 
        \subfigure[]{\includegraphics[width=0.20\linewidth]{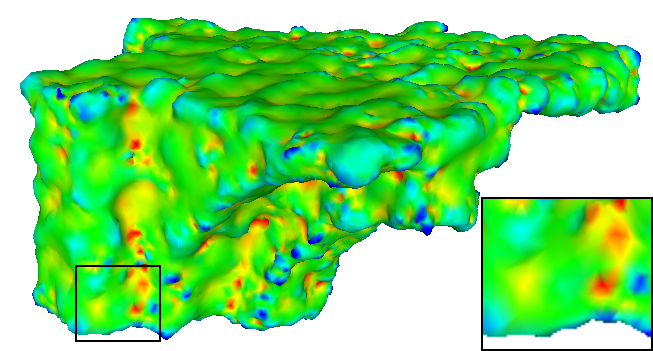}}
        \subfigure[]{\includegraphics[width=0.20\linewidth]{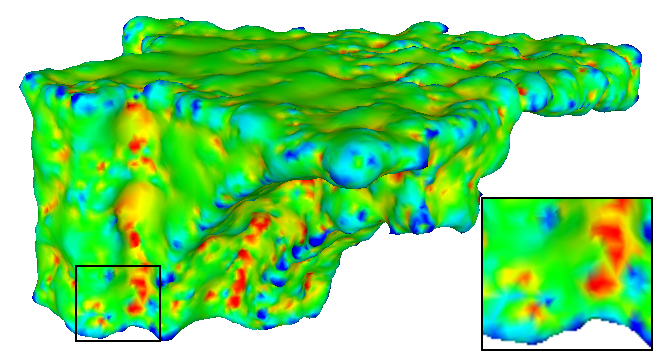}}
        \subfigure[]{\includegraphics[width=0.20\linewidth]{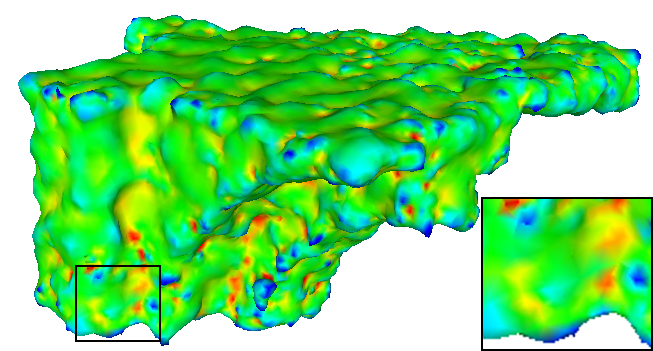}}  
        \subfigure[]{\includegraphics[width=0.20\linewidth]{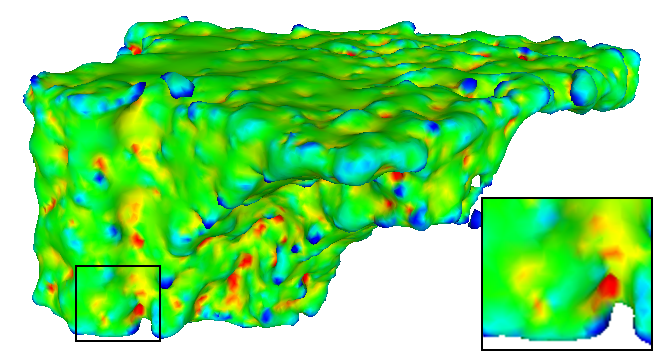}}\\ 
        \subfigure[]{\includegraphics[width=0.20\linewidth]{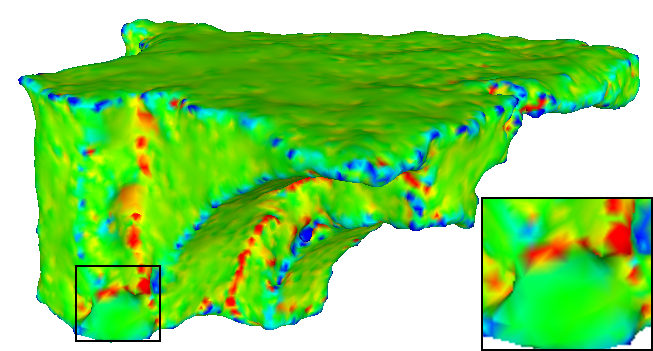}} 
        \subfigure[]{\includegraphics[width=0.20\linewidth]{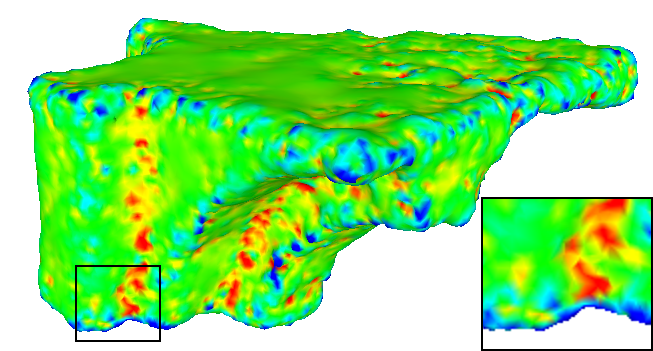}}
        \subfigure[]{\includegraphics[width=0.20\linewidth]{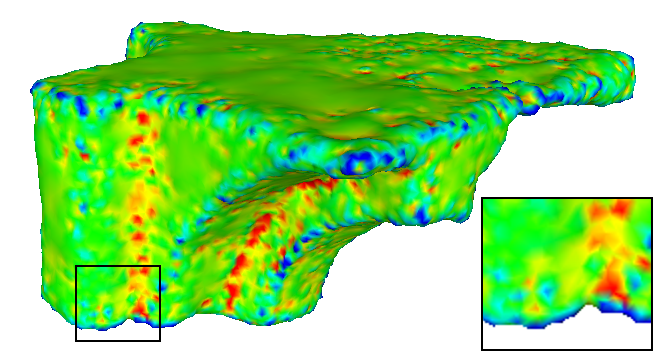}}
        \subfigure[]{\includegraphics[width=0.20\linewidth]{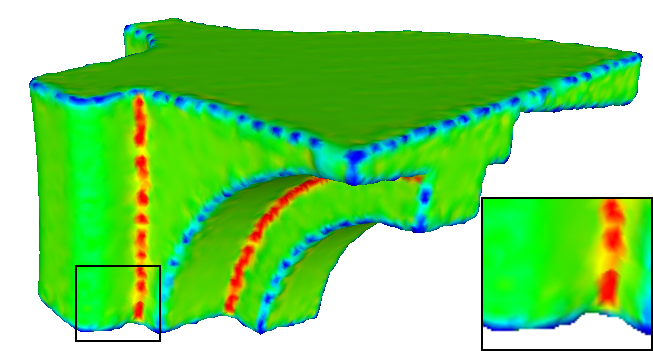}}
\caption{\textit{Fandisk} model ($\sigma=0.02$) illustration. Surface reconstruction with (a) noisy input, denoising results of (b) APSS, (c) AWLOP, (d) NLD, (e) MRPCA, (f) LR, (g) proposed GLR, and (h) ground truth, colorized by mean curvature.}
\label{fig:fan}
\end{figure*}

Denoising results of the \textit{Fandisk} model are shown in Fig.\,\ref{fig:fan} with noise level $\sigma=0.02$. The corner part is highlighted by a black rectangle, enlarged and placed at the lower-right corner.
APSS result is over-smoothed, AWLOP and NLD do not show competitive results, and MRPCA generates extra surface as shown in the black rectangle.
The patch-based LR and GLR is visually better, without over-smoothing or extra feature generated.

% llustrations of the \textit{Lordquas} model are shown in Fig.\,\ref{fig:lordquas}, with a higher noise level $\sigma=0.04$.
% In particular, the cigarette part is marked by a black rectangle, enlarged and placed at the upper-left corner.
% APSS and RIMLS again produce over-smoothed results, as seen from the smoothed edges of the book holding in the hand. RIMLS results are inflated where the nose is enlarged.
% AWLOP and NLD are under-smoothed in the plane area, and the details are lost. For example, the cigarette part is broken due to over-smoothing. 
% MRPCA shows competing results, but the plane surface is under-smoothed and the cigarette structure is curled.
% The proposed GLR better preserves the cigarette, and structural details are preserved without over-smoothing. 

% \begin{figure*}[t]
% \centering 
%         \subfigure[]{\includegraphics[width=0.20\linewidth]{eps/lordquas_004_ny_show.png}}
%         \subfigure[]{\includegraphics[width=0.20\linewidth]{eps/lordquas_004_apss_show.png}}
%         \subfigure[]{\includegraphics[width=0.20\linewidth]{eps/lordquas_004_rimls_show.png}}  
%         \subfigure[]{\includegraphics[width=0.20\linewidth]{eps/lordquas_004_awlop_show.png}}\\ 
%         \subfigure[]{\includegraphics[width=0.20\linewidth]{eps/lordquas_004_nld_show.png}} 
%         \subfigure[]{\includegraphics[width=0.20\linewidth]{eps/lordquas_004_mrpca_show.png}}
%         \subfigure[]{\includegraphics[width=0.20\linewidth]{eps/lordquas_004_glr_show.png}}
%         \subfigure[]{\includegraphics[width=0.20\linewidth]{eps/lordquas_004_gt_show.png}}
% \caption{\textit{Lordquas} model ($\sigma=0.04$) illustration. Surface reconstruction with (a) noisy input, denoising results of (b) APSS, (c) RIMLS, (d) AWLOP, (e) NLD, (f) MRPCA, (g) proposed GLR, and (h) ground truth, colorized by mean curvature.}
% \label{fig:lordquas}
% \end{figure*}

\begin{figure}[t]
\centering
    % \subfigure[]{\includegraphics[width=0.20\linewidth]{eps/anchor_gt_color08.png}}
    \subfigure[]{\includegraphics[width=0.40\linewidth]{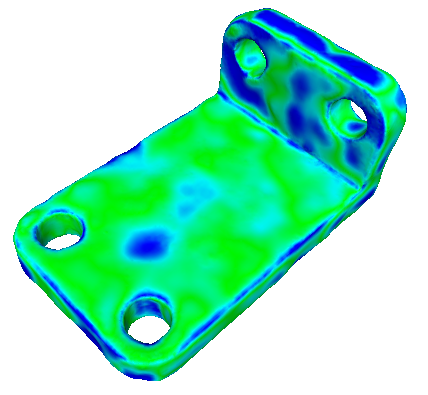}}  
    \subfigure[]{\includegraphics[width=0.40\linewidth]{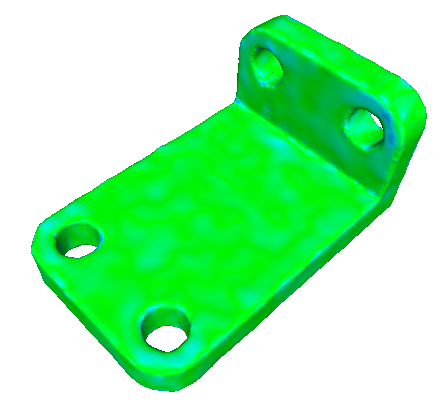}} 
\caption{\blue{Surface reconstruction of denoising results of (a) normalized Laplacian regularization and (b) combinatorial Laplacian regularization, where the surface is colorized by the distance from the ground truth surface (green means zero error, while blue means large error).}}
\label{fig:norm}
\end{figure}

\blue{We further compare GLR approach using combinatorial and normalized Laplacian matrix for the regularization. As discussed in Section \ref{sec:normalized}, normalized Laplacian cannot handle constant signal, \textit{e.g.}, a flat surface. This is consistent with the visual comparison in Fig.\;\ref{fig:norm}, where the surface reconstruction of the resulting point cloud is colorized by distance from the ground truth surface. 
Normalized Laplacian cannot even denoise a flat surface with obvious error (colored in blue), while combinatorial Laplacian preserves both the smooth surface and the sharp edges.
For numerical evaluation in term of MSE, combinatorial Laplacian outperforms normalized Laplacian by 0.014, 0.020, 0.080 for $\sigma=0.02,0.03,0.04$.}

\subsection{Generalization to ShapeNetCore Dataset}
\blue{We now test the parameter setting learned in Section \ref{sec:optpara} and \ref{sec:optpara2} with ShapeNetCore dataset \cite{shapenet2015}.
ShapeNetCore dataset is a subset of ShapeNet dataset containing 55 object categories with 52491 unique 3D models. 
The 10 categories with the largest number of models are used.
We then randomly select 10 models from each category for testing, so 100 models are used in total. 
Moreover, the 3D models in ShapeNetCore dataset are low poly meshes, so we sample approximately 30000 points on each mesh to obtain the point cloud using Poisson-disk sampling \cite{corsini2012efficient}.}

\blue{We add Gaussian noise with $\sigma =$ 0.02, 0.03, 0.04 to the models, then apply different denoising methods. 
The parameter setting for GLR is the same as used in Section \ref{sec:parameter}. For APSS and RIMLS, we try each filter scale in $\{5,6,7,8,9,10\}$ and choose the one with the best result. For other methods, we empirically set the parameters as shown in Table \ref{tab:shapepara} based on the results of previous 8 models since it is too time-consuming to tune parameters for each of the 100 models. Parameters not in Table \ref{tab:shapepara} follow the default setting.}

\begin{table}[htbp]
  \centering 
  \caption{\blue{Parameter Setting of Competing Methods for ShapeNetCore Dataset with Different Noise Levels}}
    \begin{tabular}{|l|l|c|}
    \hline
   Methods & Parameters& \multicolumn{1}{c|}{$\sigma$ 0.02\,$|$\,0.03\,$|$\,0.04}  \\
    \hline
    APSS  & \multicolumn{1}{l|}{filter scale} & exhaustive search \{5,6,7,8,9,10\}   \\
    \hline
    RIMLS & \multicolumn{1}{l|}{filter scale} & exhaustive search \{5,6,7,8,9,10\}  \\
    \hline
    \multirow{2}[1]{*}{AWLOP} & \multicolumn{1}{l|}{repulsion force} & \multicolumn{1}{c|}{0.3\,$|$\,0.3\,$|$\,0.3} \\
    \cline{2-3}
          & \multicolumn{1}{l|}{iteration} & \multicolumn{1}{c|}{2\,$|$\,2\,$|$\,2}\\
    \hline
  \multirow{2}[1]{*}{MRPCA}  & \multicolumn{1}{l|}{data fitting} & \multicolumn{1}{c|}{1\,$|$\,1\,$|$\,1} \\
    \cline{2-3}
          & \multicolumn{1}{l|}{iteration} & \multicolumn{1}{c|}{6\,$|$\,6\,$|$\,6} \\
    \hline
    \end{tabular}%
  \label{tab:shapepara}%
\end{table}%

\begin{figure*}[t] % model 63
\centering
        \subfigure[]{\includegraphics[width=0.24\linewidth]{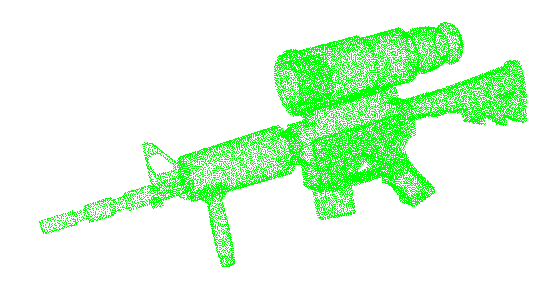}}  
        \subfigure[]{\includegraphics[width=0.24\linewidth]{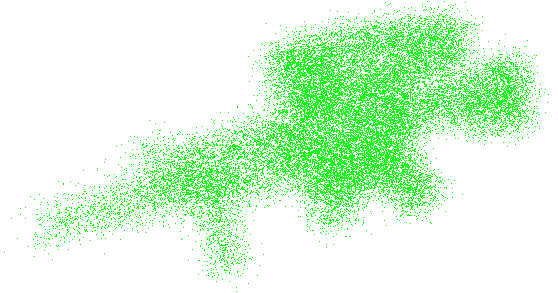}}
        \subfigure[]{\includegraphics[width=0.24\linewidth]{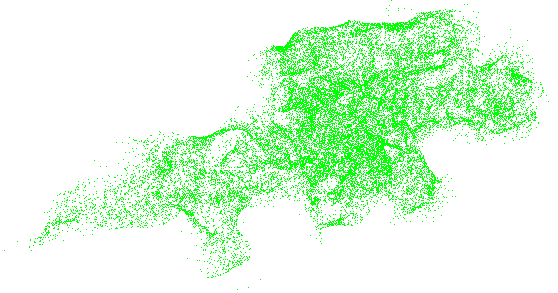}}
        % \subfigure[]{\includegraphics[width=0.30\linewidth]{eps/rimls_crop.png}}
        \subfigure[]{\includegraphics[width=0.24\linewidth]{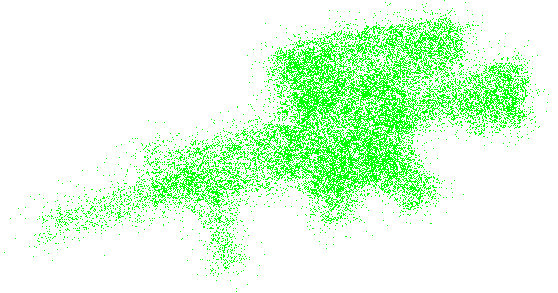}}  
        \\
        \subfigure[]{\includegraphics[width=0.24\linewidth]{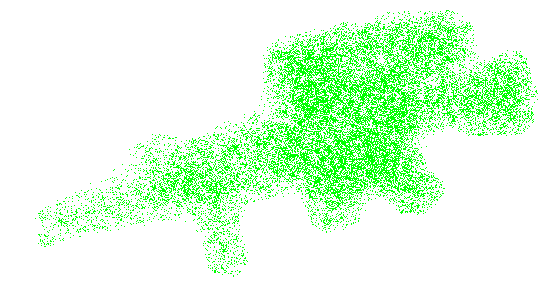}}
        \subfigure[]{\includegraphics[width=0.24\linewidth]{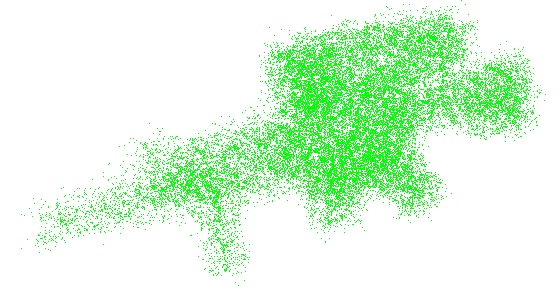}}  
        \subfigure[]{\includegraphics[width=0.24\linewidth]{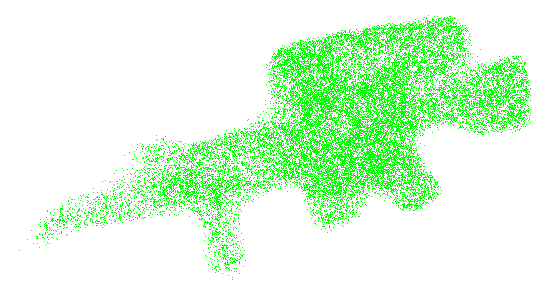}}
        \subfigure[]{\includegraphics[width=0.24\linewidth]{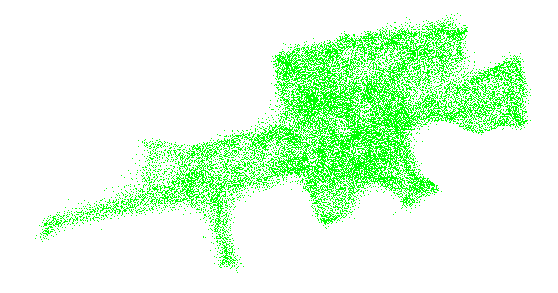}}  
\caption{\blue{Rifle model from ShapeNetCore dataset. (a) ground truth, (b) noisy input with $\sigma=0.04$, denoising results of (c) APSS, (d) AWLOP, (e) NLD, (f) MRPCA, (g) LR, (h) GLR.}}
\label{fig:lowpoly}
\end{figure*}

\blue{The MSE, SNR and MCD results are compared with competing methods in Table \ref{tab:shapenetmse}, \ref{tab:shapenetsnr} and \ref{tab:shapenetmcd} where GLR provides the best results, and the patch-based LR is the second best. 
At high noise level, \textit{e.g.}, Fig.\,\ref{fig:lowpoly}(b), the points distract from the surface and tend to fill the bulk of the object, so for methods based on plane fitting, the denoising leads to erroneous results, \textit{e.g.}, APSS in Fig.\,\ref{fig:lowpoly}(c). RIMLS provides similar results as APSS thus is not shown in Fig.\,\ref{fig:lowpoly}.
The noise in results of AWLOP and MRPCA in Fig.\,\ref{fig:lowpoly}(d) and (f) is not fully removed with noticeable outliers.
NLD in Fig.\,\ref{fig:lowpoly}(e) provides smooth results without outliers, which demonstrates the robustness of non-local means filtering against the above approaches at high noise level.
Patch-based LR and GLR in Fig.\,\ref{fig:lowpoly}(g) and (h) provide the best results, where the shape of the rifle model is well preserved, validating the effectiveness of patch-similarity based filtering.
However, LR tends to over-smooth the model and fine details are lost during patch extraction procedure, while the proposed GLR preserves the salient features without over-smoothing.
In sum, the generalization to ShapeNetCore dataset validates the robustness of parameter setting in Section \ref{sec:parameter} as well as the superiority of patch-based filtering over other approaches.}

% Different from the results on small dataset, LR provides the second best results on ShapeNetCore dataset. This is because high noise level input in ShapeNetCore does not contain fine details, so the results of high noise level is not affected by the lose of fine details during patch extraction.
% On the contrary, the smoothness induced by patch extraction and low-rank dictionary optimization benefits the denoising.

\subsection{Complexity Analysis}
The computational complexity of different algorithms are summarized in Table \ref{tab:complexity}. $N$ is the number of points, $r$ is the number of iterations of implementing the algorithm since all algorithms except NLD and LR adopt iterative restoration, $K$ is the neighborhood size chosen for different operation in different schemes. Parameters used in specific methods are explained along with the complexity. The parameter ranges in Table \ref{tab:complexity} are suggested in the original papers.

As shown in Table \ref{tab:complexity}, APSS, RIMLS and NLD have the lowest complexity. 
% Considering APSS provides the second best performance, it is a good trade-off between complexity and performance, which is also validated in \cite{han2017review}.
MRPCA and GLR are of similar complexity; MRPCA's can be higher due to large $K$ and $t$.
\blue{The complexity of LR is high due to complexity in solving low-rank matrix factorization for dictionary learning.}
GLR have relatively high complexity, but provides the best performance as shown in the above evaluation, so GLR is favorable if the requirement for denoising accuracy is high.

\blue{We additional include the runtime of different methods implemented on Intel i7-8550U CPU at 1.80GHz and 8GB RAM. Since the methods are implemented with different programming language and C++ is known to far surpass the speed of Matlab \cite{andrews2012computation}, we cannot directly use the runtime for complexity comparison. Nevertheless, the runtime of NLD is more than 10 times that of APSS while the complexity is approximately the same as APSS, thus if implemented in C++, the runtime of GLR can be reduced by 10 times potentially.}

\begin{table}[t]
\caption{Time Complexity Summary of Different Schemes}
\centering
\begin{tabular}{c|l}
\hline   Method  & Complexity \\ \hline \hline
APSS &  $O(rKN\log N)$, $r \le 15$, $K \in (16, 100)$\\ \hline
RIMLS &  Same as above \\ \hline　
AWLOP & $O(r(\sigma_p N^2 + kN\log N))$, $r\approx3$\\
      & $\sigma_p$ for neighborhood radius \\
      &  $k=6$ for PCA-based normal estimation\\\hline
NLD   &  $ O(N+2KN\log N)$, $K \approx 20$ \\ \hline % K=200
MRPCA &  $ O(r(KN\log N + N\log N + tKN))$ \\
      & $r\le20$, $K\in(30,100)$,\\
      & $t \in (50,100)$ is the RPCA solver iteration number\\ \hline
\blue{LR}   &  $\mathcal{O}(l/\tau(h^2+M))$, $l=50$ is dictionary atom number \\
    & $h^2=16^2$ is patch grid size, $M$ is patch number\\
    & $\tau = 10^{-5}$ is proximal gradient descent step size \\\hline
GLR &    $ O(rKM(\log M + k\log k + k\sqrt{C}))$, $r\le15$,\\
    &    $M=N/2$, $K=16$, $k=30$,  $C\le1+2/\mu \le 1.921$ \\
    &    ($\mu \ge 25(\exp(1/12)-1) \approx 2.173$) \\ \hline
\end{tabular}
\label{tab:complexity}
\end{table}

\begin{table}[htbp]
  \centering \scriptsize
  \caption{\blue{Average Runtime (sec) on ShapeNetCore Dataset and Programming Language for Different Methods}}
    \begin{tabular}{|r|r|r|r|r|r|r|}
    \hline \multicolumn{1}{|r|}{APSS} & \multicolumn{1}{r|}{RIMLS} & \multicolumn{1}{r|}{AWLOP} & \multicolumn{1}{r|}{NLD} & \multicolumn{1}{r|}{MRPCA} & \multicolumn{1}{r|}{LR} & \multicolumn{1}{r|}{GLR}  \\
    \hline C++ & C++ & C++ & Matlab & C++ & Matlab & Matlab  \\
    \hline 13.8 & 18.1 & 21.2 & 156.4 & 18.0 & 464.7 & 372.2  \\
    \hline
    \end{tabular}%
  \label{tab:shapenetrt}%
\end{table}%

\begin{table*}[thb]
\centering
\caption{SNR (dB) Results of Different Methods on Small Dataset with Three Noise Levels}
\begin{tabular}{|c|c|c|c|c|c|c|c|c|c|c|} 
\hline
Noise level         & Methods & Anchor         & Bimba  & Bunny  & Daratech       & DC     & Fandisk        & Gargoyle       & Lordquas       & Average  \\ 
\hline\hline
\multirow{8}{*}{$\sigma$ = 0.02} & Noisy   & 47.41  & 41.40  & 51.99  & 45.85  & 46.42  & 34.06  & 46.91  & 46.61  &   45.08       \\ 
\cline{2-11}
   & APSS    & 49.61  & 45.13  & 54.20  & 47.71  & 48.83  & 36.52  & 49.01  & 49.27  &    47.53      \\ 
\cline{2-11}
   & RIMLS   & 49.41  & 42.60  & 53.70  & 47.44  & 48.23  & 36.80  & 48.57  & 48.60  &  46.92        \\ 
\cline{2-11}
   & AWLOP   & 48.31  & \textbf{46.91}  & 52.98  & 46.56  & 47.59  & 37.06  & 48.01  & 47.92  &  46.92        \\ 
\cline{2-11}
   & NLD     & 48.53  & 42.30  & 53.14  & 46.82  & 47.82  & 36.16  & 48.01  & 48.22  &   46.38       \\ 
\cline{2-11}
   & MRPCA   & 49.88  & 43.53  & 53.45  & 46.72  & 48.68  & \textbf{38.55} & 48.66  & 49.27  &    47.34      \\ 
\cline{2-11}
   & LD   & 48.69 & 45.06 & 53.15 & 47.27 & 47.83 & 38.00 & 47.57 & 48.78 & 47.04      \\ 
% \cline{2-11}
%   & GLR     & \textbf{50.52} & \textbf{48.18} & \textbf{54.76} & \textbf{47.80} & \textbf{49.22} & 37.99  & \textbf{49.18} & \textbf{49.69} &  \textbf{48.42}        \\ 
\cline{2-11}
   & GLR     & \textbf{50.55} & 46.00 & \textbf{54.95} & \textbf{48.02} & \textbf{49.34} & 38.05 & \textbf{49.30} & \textbf{49.81} & \textbf{48.25}         \\ 
\hline\hline
\multirow{8}{*}{$\sigma$ = 0.03} & Noisy   & 45.25  & 38.42  & 49.72  & 43.70  & 44.32  & 31.75  & 44.75  & 44.58  &   42.81       \\ 
\cline{2-11}
   & APSS    & 48.24  & 41.17  & 52.77  & 46.00  & 47.64  & 35.07  & 47.63  & 48.34  &     45.86     \\ 
\cline{2-11}
   & RIMLS   & 48.00  & 40.36  & 52.22  & 45.46  & 46.94  & 34.38  & 47.12  & 47.57  &  45.26        \\ 
\cline{2-11}
   & AWLOP   & 46.69  & \textbf{45.02} & 51.24  & 45.12  & 46.04  & 35.71  & 46.39  & 46.53  & 45.34         \\ 
\cline{2-11}
   & NLD     & 47.16  & 38.91  & 51.67  & 45.34  & 46.49  & 33.07  & 46.68  & 46.89  &     44.53     \\ 
\cline{2-11}
   & MRPCA   & 48.60  & 39.40  & 52.32  & 45.18  & 47.62  & 34.83  & 47.52  & 48.40  &   45.48       \\ 
\cline{2-11}
   & LD   & 47.91 & 40.52 & 52.38 & 45.59 & 47.10 & 36.12 & 46.88 & 48.09 & 45.57      \\ 
% \cline{2-11}
%   & GLR     & \textbf{49.09} & 44.60  & \textbf{53.27} & \textbf{46.13} & \textbf{48.02} & \textbf{36.10} & \textbf{47.88} & \textbf{48.94} &   \textbf{46.75 }      \\ 
\cline{2-11}
   & GLR     & \textbf{49.20} & 44.03 & \textbf{53.28} & \textbf{46.13} &\textbf{ 47.94} & \textbf{37.09} & \textbf{47.87} & \textbf{49.00} & \textbf{46.82}     \\ 
\hline\hline
\multirow{8}{*}{$\sigma$ = 0.04} & Noisy   & 43.78  & 36.13  & 48.31  & 42.34  & 42.86  & 29.95  & 43.31  & 43.09  &  41.22        \\ 
\cline{2-11}
   & APSS    & 47.60  & 40.94  & 52.09  & 44.46  & \textbf{46.84}  & 33.02  & 46.69  & 47.68  &   44.92       \\ 
\cline{2-11}
   & RIMLS   & 47.27  & 38.76  & 51.22  & 43.58  & 45.71  & 32.23  & 46.14  & 46.80  &   43.96       \\ 
\cline{2-11}
   & AWLOP   & 45.74  & \textbf{43.73} & 50.32  & 44.32  & 45.11  & 34.77 & 45.44  & 46.85  & 44.54         \\ 
\cline{2-11}
   & NLD     & 46.02  & 36.39  & 50.54  & 43.98  & 45.15  & 30.44  & 45.53  & 45.40  &   42.93       \\ 
\cline{2-11}
   & MRPCA   & 48.09  & 36.71  & 51.93  & 44.25  & 47.00  & 31.19  & 46.88  & 47.80  &   44.23       \\ 
\cline{2-11}
   & LD     & 47.41 & 36.50 & 51.89 & 44.41 & 46.54 & 32.68 & 46.44 & 47.52 & 44.17       \\ 
% \cline{2-11}
%   & GLR     & \textbf{48.11} & 41.15  & \textbf{52.46} & \textbf{44.88} & \textbf{47.10} & 34.70  & \textbf{47.01} & \textbf{48.45} &   \textbf{45.48}       \\
\cline{2-11}
   & GLR     & \textbf{48.67} & 42.22 & \textbf{52.51} & \textbf{44.64} & 46.80 & \textbf{35.20} & \textbf{46.89} & \textbf{48.40} & \textbf{45.67}       \\
\hline
\end{tabular}
\label{tab:modelSNR}
\end{table*}

\begin{table*}[thb]
\centering
\caption{MCD Results of Different Methods on Small Dataset with Three Noise Levels}
\begin{tabular}{|c|c|c|c|c|c|c|c|c|c|c|} 
\hline
Noise level      & Methods & Anchor & Bimba   & Bunny  & Daratech        & DC     & Fandisk & Gargoyle        & Lordquas        & Average \\ \hline \hline
\multirow{8}{*}{$\sigma$ = 0.02} & Noisy   & 0.384  & 0.0268  & 0.366  & 0.364  & 0.350  & 0.0368  & 0.380  & 0.331  & 0.280   \\ \cline{2-11} 
        & APSS    & 0.302  & 0.0188  & 0.293  & 0.300  & 0.275  & 0.0294  & 0.308  & 0.252  & 0.222   \\ \cline{2-11} 
        & RIMLS   & 0.311  & 0.0240  & 0.309  & 0.310  & 0.292  & 0.0287  & 0.322  & 0.271  & 0.233   \\ \cline{2-11} 
        & AWLOP   & 0.353  & \textbf{0.0161}  & 0.332  & 0.339  & 0.313  & 0.0280  & 0.342  & 0.292  & 0.252   \\ \cline{2-11} 
        & NLD     & 0.339  & 0.0247  & 0.325  & 0.327  & 0.304  & 0.0304  & 0.340  & 0.281  & 0.246   \\ \cline{2-11} 
        & MRPCA   & 0.289  & 0.0219  & 0.316  & 0.331  & 0.278  & \textbf{0.0239} & 0.319  & 0.251  & 0.229   \\ \cline{2-11}
        & LD   & 0.330 & 0.0190 & 0.326 & 0.313 & 0.303 & 0.0254 & 0.355 & 0.264 & 0.242 \\ 
        % \cline{2-11} 
        % & GLR     & \textbf{0.271} & \textbf{0.0141} & \textbf{0.277} & \textbf{0.297} & \textbf{0.264} & 0.0253  & \textbf{0.303} & \textbf{0.241} & \textbf{0.212 }  \\ 
        \cline{2-11} 
        & GLR     & \textbf{0.272} & 0.0174 & \textbf{0.272} & \textbf{0.290} & \textbf{0.261} & 0.0252 & \textbf{0.299} &\textbf{ 0.238} & \textbf{0.209} \\
\hline\hline
\multirow{8}{*}{$\sigma$ = 0.03} & Noisy   & 0.475  & 0.0356  & 0.456  & 0.449  & 0.430  & 0.0453  & 0.469  & 0.402  & 0.345   \\ \cline{2-11} 
        & APSS    & 0.348  & 0.0274  & 0.338  & 0.358  & 0.309  & 0.0338  & 0.353  & 0.277  & 0.255   \\ \cline{2-11} 
        & RIMLS   & 0.360  & 0.0297  & 0.357  & 0.379  & 0.333  & 0.0361  & 0.372  & 0.300  & 0.271   \\ \cline{2-11} 
        & AWLOP   & 0.415  & \textbf{0.0194} & 0.395  & 0.392  & 0.365  & 0.0320  & 0.401  & 0.335  & 0.294   \\ \cline{2-11} 
        & NLD     & 0.391  & 0.0341  & 0.377  & 0.380  & 0.347  & 0.0405  & 0.388  & 0.321  & 0.285   \\ \cline{2-11} 
        & MRPCA   & 0.331  & 0.0325  & 0.353  & 0.386  & 0.310  & 0.0343  & 0.356  & 0.274  & 0.260   \\ \cline{2-11}
        & LD   & 0.359 & 0.0293 & 0.352 & 0.372 & 0.326 & 0.0304 & 0.379 & 0.284 & 0.266   \\
        % \cline{2-11} 
        % & GLR     & \textbf{0.316} & 0.0198  &  \textbf{0.322} & \textbf{0.352} &\textbf{0.298} & \textbf{0.0303} & \textbf{0.344} & \textbf{0.261} & \textbf{0.243}   \\ 
        \cline{2-11} 
        & GLR     & \textbf{0.312} & 0.0209 & \textbf{0.322} & \textbf{0.353} & \textbf{0.300} &\textbf{ 0.0276} & \textbf{0.345} & \textbf{0.259} & \textbf{0.242} \\
\hline\hline
\multirow{8}{*}{$\sigma$ = 0.04} & Noisy   & 0.545  & 0.0445  & 0.521  & 0.510  & 0.494  & 0.0535  & 0.539  & 0.462  & 0.396   \\ \cline{2-11} 
        & APSS    & 0.375  & 0.0278  & 0.362  & 0.417  & \textbf{0.336}  & 0.0409  & 0.388  & 0.297  & 0.280   \\ \cline{2-11} 
        & RIMLS   & 0.389  & 0.0340  & 0.395  & 0.454  & 0.376  & 0.0442  & 0.410  & 0.325  & 0.303   \\ \cline{2-11} 
        & AWLOP   & 0.456  & \textbf{0.0220} & 0.432  & 0.425  & 0.400  & 0.0351  & 0.441  & 0.322  & 0.317   \\ \cline{2-11} 
        & NLD     & 0.439  & 0.0433  & 0.421  & 0.435  & 0.397  & 0.0514  & 0.434  & 0.372  & 0.324   \\ \cline{2-11} 
        & MRPCA   & 0.351 & 0.0421  & 0.367  & 0.424  & 0.330  & 0.0479  & 0.380  & 0.293  & 0.279   \\ \cline{2-11} 
        & LD&   0.380 & 0.0430 & 0.369 & 0.418 & 0.345 & 0.0419 & 0.397 & 0.302 & 0.287 \\ 
        % \cline{2-11} 
        % & GLR     & 0.353  & 0.0274  & \textbf{0.349} & \textbf{0.400} & \textbf{0.327} & \textbf{0.0345} & \textbf{0.376} & \textbf{0.275} &\textbf{ 0.268}   \\ 
        \cline{2-11} 
        & GLR &  \textbf{ 0.334} & 0.0248 & \textbf{0.347} & \textbf{0.411} & 0.337 & \textbf{0.0330} & \textbf{0.379} & \textbf{0.277} & \textbf{0.268} \\
        \hline
\end{tabular}
\label{tab:modell1}
\end{table*}

\begin{table*}[htbp]
  \centering
  \caption{\blue{MSE ($\times 10^{-3}$) Results of Different Methods for 10 Categories of ShapeNetCore Dataset}}
    \begin{tabular}{|c|l|r|r|r|r|r|r|r|r|r|r|r|}
    \hline
    \multicolumn{1}{|l|}{Noise Level} & Methods & \multicolumn{1}{l|}{plane} & \multicolumn{1}{l|}{bench} & \multicolumn{1}{l|}{car} & \multicolumn{1}{l|}{chair} & \multicolumn{1}{l|}{lamp} & \multicolumn{1}{l|}{speaker} & \multicolumn{1}{l|}{rifle} & \multicolumn{1}{l|}{sofa} & \multicolumn{1}{l|}{table} & \multicolumn{1}{l|}{vessel} & \multicolumn{1}{l|}{Average}   \\
    \hline
    \hline
    \multirow{8}[0]{*}{0.02} & Noisy & 4.988 & 6.206 & 6.188 & 6.709 & 5.509 & 6.656 & 4.911 & 6.919 & 6.283 & 5.605 & 5.997   \\
\cline{2-13}          & APSS  & 4.059 & 4.783 & 4.380 & 4.693 & 4.032 & \textbf{3.643} & 5.066 & 4.370 & 5.204 & 4.060 & 4.429   \\
\cline{2-13}          & RIMLS & 4.718 & 5.479 & 4.904 & 5.446 & 4.873 & 4.134 & 5.236 & 4.838 & 5.817 & 4.927 & 5.037   \\
\cline{2-13}          & AWLOP & 3.890 & 5.046 & 5.196 & 5.690 & 4.034 & 5.424 & 3.646 & 5.951 & 5.237 & 4.342 & 4.846   \\
\cline{2-13}          & NLD   & 4.347 & 5.096 & 4.902 & 5.247 & 4.700 & 5.091 & 4.505 & 5.192 & 5.041 & 4.819 & 4.894   \\
\cline{2-13}          & MRPCA & 4.413 & 5.126 & 4.915 & 5.073 & 4.670 & 4.772 & 4.560 & 5.095 & 5.229 & 4.818 & 4.867   \\
\cline{2-13}          & LR    & \textbf{3.434} & 4.802 & 4.389 & 5.159 & \textbf{3.537} & 3.689 & \textbf{3.199} & 4.760 & 4.756 & \textbf{3.734} & 4.146   \\
\cline{2-13}          & GLR   & 3.489 & \textbf{4.368} & \textbf{4.067} & \textbf{4.355} & 3.714 & 3.752 & 3.640 & \textbf{4.294} & \textbf{4.565} & 3.789 & \textbf{4.003}   \\
    \hline
    \hline
    \multirow{8}[0]{*}{0.03} & Noisy & 6.393 & 7.848 & 7.829 & 8.540 & 7.302 & 8.707 & 6.627 & 8.858 & 7.933 & 7.210 & 7.725   \\
\cline{2-13}          & APSS  & 6.354 & 6.687 & 5.636 & 6.377 & 5.705 & 4.886 & 8.149 & 5.832 & 6.814 & 6.048 & 6.249   \\
\cline{2-13}          & RIMLS & 7.237 & 7.441 & 6.907 & 7.963 & 7.454 & 6.256 & 7.579 & 6.886 & 7.585 & 7.592 & 7.290   \\
\cline{2-13}          & AWLOP & 5.227 & 6.899 & 7.108 & 7.824 & 5.762 & 7.779 & 4.926 & 8.242 & 6.942 & 6.076 & 6.679   \\
\cline{2-13}          & NLD   & 5.946 & 7.091 & 6.884 & 7.351 & 6.678 & 7.459 & 6.334 & 7.584 & 6.963 & 6.641 & 6.893   \\
\cline{2-13}          & MRPCA & 6.034 & 7.082 & 6.873 & 7.211 & 6.649 & 7.145 & 6.403 & 7.409 & 7.099 & 6.709 & 6.861   \\
\cline{2-13}          & LR    & \textbf{4.229} & 5.252 & \textbf{4.638} & 5.498 & 4.636 & \textbf{4.357} & 5.144 & \textbf{5.412} & 5.495 & \textbf{4.545} & 4.920   \\
\cline{2-13}          & GLR   & 4.274 & \textbf{5.234} & 4.808 & \textbf{5.496} & \textbf{4.518} & 4.709 & \textbf{4.553} & 5.452 & \textbf{5.297} & 4.650 & \textbf{4.899}   \\
    \hline
    \hline
    \multirow{8}[0]{*}{0.04} & Noisy & 7.784 & 9.433 & 9.443 & 10.259 & 9.127 & 10.698 & 8.443 & 10.650 & 9.619 & 8.790 & 9.425   \\
\cline{2-13}          & APSS  & 9.020 & 8.626 & 7.474 & 9.173 & 8.457 & 6.943 & 10.083 & 7.983 & 9.031 & 9.533 & 8.632   \\
\cline{2-13}          & RIMLS & 9.073 & 10.179 & 9.084 & 10.545 & 9.790 & 8.836 & 9.921 & 10.268 & 10.070 & 9.894 & 9.766   \\
\cline{2-13}          & AWLOP & 6.757 & 8.748 & 8.948 & 9.807 & 7.833 & 10.100 & 6.700 & 10.307 & 8.864 & 7.953 & 8.602   \\
\cline{2-13}          & NLD   & 7.431 & 8.856 & 8.773 & 9.378 & 8.640 & 9.788 & 8.182 & 9.734 & 8.885 & 8.391 & 8.806   \\
\cline{2-13}          & MRPCA & 7.525 & 8.858 & 8.786 & 9.288 & 8.612 & 9.520 & 8.245 & 9.622 & 8.961 & 8.448 & 8.786   \\
\cline{2-13}          & LR    & 5.799 & \textbf{6.244} & \textbf{5.550} & \textbf{6.469} & 6.441 & \textbf{5.638} & 7.456 & \textbf{6.396} & 6.229 & 6.209 & 6.243   \\
\cline{2-13}          & GLR   & \textbf{5.320} & 6.432 & 6.029 & 6.746 & \textbf{5.809} & 5.993 & \textbf{5.494} & 6.801 & \textbf{6.025} & \textbf{5.880} & \textbf{6.053}   \\
    \hline
    \end{tabular}%
  \label{tab:shapenetmse}%
\end{table*}%

\begin{table*}[htbp]
  \centering
  \caption{\blue{SNR (dB) Results of Different Methods for 10 Categories of ShapeNetCore Dataset}}
    \begin{tabular}{|c|l|r|r|r|r|r|r|r|r|r|r|r|}
    \hline
    \multicolumn{1}{|l|}{Noise Level} & Methods & \multicolumn{1}{l|}{plane} & \multicolumn{1}{l|}{bench} & \multicolumn{1}{l|}{car} & \multicolumn{1}{l|}{chair} & \multicolumn{1}{l|}{lamp} & \multicolumn{1}{l|}{speaker} & \multicolumn{1}{l|}{rifle} & \multicolumn{1}{l|}{sofa} & \multicolumn{1}{l|}{table} & \multicolumn{1}{l|}{vessel} & \multicolumn{1}{l|}{Average}   \\
    \hline
    \hline
    \multirow{8}[0]{*}{0.02} & Noisy & 36.58 & 38.50 & 38.40 & 37.42 & 40.95 & 38.62 & 38.06 & 37.70 & 39.38 & 37.59 & 38.32\\
\cline{2-13}          & APSS  & 38.71 & 41.23 & 41.92 & 41.04 & 44.19 & \textbf{44.66} & 37.87 & 42.40 & 41.29 & 40.94 & 41.42   \\
\cline{2-13}          & RIMLS & 37.16 & 39.90 & 40.78 & 39.55 & 42.22 & 43.39 & 37.49 & 41.37 & 40.17 & 39.01 & 40.10   \\
\cline{2-13}          & AWLOP & 39.06 & 40.60 & 40.15 & 39.14 & 44.14 & 40.75 & 41.04 & 39.23 & 41.26 & 40.13 & 40.55   \\
\cline{2-13}          & NLD   & 37.93 & 40.45 & 40.72 & 39.85 & 42.51 & 41.28 & 38.90 & 40.55 & 41.55 & 39.08 & 40.28   \\
\cline{2-13}          & MRPCA & 37.78 & 40.40 & 40.69 & 40.19 & 42.56 & 41.92 & 38.77 & 40.75 & 41.18 & 39.08 & 40.33   \\
\cline{2-13}          & LR    & \textbf{40.43} & 41.21 & 41.92 & 40.16 & \textbf{45.61} & 44.59 & \textbf{42.31} & 41.57 & 42.22 & \textbf{41.88} & 42.19   \\
\cline{2-13}          & GLR   & 40.15 & \textbf{42.06} & \textbf{42.61} & \textbf{41.75} & 44.86 & 44.33 & 41.01 & \textbf{42.51} & \textbf{42.61} & 41.50 & \textbf{42.34}   \\
    \hline
    \hline
    \multirow{8}[0]{*}{0.03} & Noisy & 34.13 & 36.17 & 36.06 & 35.02 & 38.14 & 35.94 & 35.10 & 35.24 & 37.05 & 35.11 & 35.80  \\
\cline{2-13}          & APSS  & 34.27 & 37.97 & 39.42 & 37.99 & 40.73 & 41.76 & 33.26 & 39.53 & 38.60 & 37.02 & 38.05   \\
\cline{2-13}          & RIMLS & 33.03 & 36.82 & 37.43 & 35.93 & 38.08 & 39.47 & 33.81 & 37.87 & 37.62 & 34.73 & 36.48   \\
\cline{2-13}          & AWLOP & 36.14 & 37.47 & 37.02 & 35.93 & 40.58 & 37.11 & 38.11 & 35.96 & 38.46 & 36.79 & 37.36   \\
\cline{2-13}          & NLD   & 34.83 & 37.17 & 37.34 & 36.49 & 39.02 & 37.45 & 35.53 & 36.78 & 38.34 & 35.92 & 36.89   \\
\cline{2-13}          & MRPCA & 34.69 & 37.18 & 37.35 & 36.68 & 39.06 & 37.88 & 35.43 & 37.01 & 38.14 & 35.82 & 36.92   \\
\cline{2-13}          & LR    & \textbf{38.26} & \textbf{40.29} & \textbf{41.34} & \textbf{39.46} & 42.70 & \textbf{42.86} & 37.64 & \textbf{40.30} & 40.78 & \textbf{39.73} & \textbf{40.33}   \\
\cline{2-13}          & GLR   & 38.12 & 40.23 & 40.92 & 39.39 & \textbf{42.90} & 42.04 & \textbf{38.79} & 40.10 & \textbf{41.11} & 39.43 & 40.30   \\
    \hline
    \hline
    \multirow{8}[0]{*}{0.04} & Noisy &32.20 & 34.35 & 34.21 & 33.22 & 35.93 & 33.90 & 32.73 & 33.42 & 35.14 & 33.18 & 33.83\\
\cline{2-13}          & APSS  & 30.95 & 35.43 & 36.68 & 34.46 & 36.88 & 38.51 & 31.12 & 36.42 & 35.86 & 32.58 & 34.89   \\
\cline{2-13}          & RIMLS & 30.78 & 33.69 & 34.76 & 33.11 & 35.32 & 36.05 & 31.18 & 33.92 & 34.87 & 32.10 & 33.58   \\
\cline{2-13}          & AWLOP & 33.59 & 35.10 & 34.74 & 33.68 & 37.50 & 34.49 & 35.02 & 33.74 & 35.98 & 34.15 & 34.80   \\
\cline{2-13}          & NLD   & 32.64 & 34.97 & 34.93 & 34.08 & 36.47 & 34.76 & 33.03 & 34.30 & 35.92 & 33.63 & 34.47   \\
\cline{2-13}          & MRPCA & 32.52 & 34.97 & 34.92 & 34.18 & 36.50 & 35.03 & 32.95 & 34.42 & 35.83 & 33.56 & 34.49   \\
\cline{2-13}          & LR    & 35.10 & \textbf{38.47} & \textbf{39.52} & \textbf{37.79} & 39.51 & \textbf{40.33} & 33.99 & 38.53 & 39.46 & 36.66 & 37.94   \\
\cline{2-13}          & GLR   & \textbf{35.81} & 38.14 & 38.58 & 37.28 & \textbf{40.38} & 39.58 & \textbf{36.89} & \textbf{37.80} & \textbf{39.78} & \textbf{37.00} & \textbf{38.12}   \\
    \hline
    \end{tabular}%
  \label{tab:shapenetsnr}%
\end{table*}%

\begin{table*}[htbp]
  \centering
  \caption{\blue{MCD ($\times 10^{-3}$) Results of Different Methods for 10 Categories of ShapeNetCore Dataset}}
    \begin{tabular}{|c|l|r|r|r|r|r|r|r|r|r|r|r|}
    \hline
    \multicolumn{1}{|l|}{Noise Level} & Methods & \multicolumn{1}{l|}{plane} & \multicolumn{1}{l|}{bench} & \multicolumn{1}{l|}{car} & \multicolumn{1}{l|}{chair} & \multicolumn{1}{l|}{lamp} & \multicolumn{1}{l|}{speaker} & \multicolumn{1}{l|}{rifle} & \multicolumn{1}{l|}{sofa} & \multicolumn{1}{l|}{table} & \multicolumn{1}{l|}{vessel} & \multicolumn{1}{l|}{Average}   \\
    \hline
    \hline
    \multirow{8}[0]{*}{0.02} & Noisy &6.84  & 8.54  & 8.63  & 9.28  & 7.54  & 9.16  & 6.51  & 9.58  & 8.69  & 7.65  & 8.24  \\
\cline{2-13}          & APSS  & 5.73  & 6.83  & 6.35  & 6.78  & 5.69  & \textbf{5.34} & 6.77  & 6.40  & 7.42  & 5.78  & 6.31   \\
\cline{2-13}          & RIMLS & 6.56  & 7.71  & 7.06  & 7.77  & 6.76  & 6.02  & 6.98  & 7.05  & 8.19  & 6.89  & 7.10   \\
\cline{2-13}          & AWLOP & 5.51  & 7.15  & 7.43  & 8.07  & 5.71  & 7.70  & 4.99  & 8.45  & 7.45  & 6.14  & 6.86   \\
\cline{2-13}          & NLD   & 6.09  & 7.24  & 7.07  & 7.52  & 6.56  & 7.28  & 6.05  & 7.51  & 7.21  & 6.74  & 6.93   \\
\cline{2-13}          & MRPCA & 6.16  & 7.25  & 7.06  & 7.27  & 6.51  & 6.84  & 6.11  & 7.36  & 7.43  & 6.72  & 6.87   \\
\cline{2-13}          & LR    & \textbf{4.92} & 6.84  & 6.34  & 7.35  & \textbf{5.05} & 5.36  & \textbf{4.45} & 6.89  & 6.82  & \textbf{5.35} & 5.94   \\
\cline{2-13}          & GLR   & 4.98  & \textbf{6.26} & \textbf{5.93} & \textbf{6.31} & 5.27  & 5.47  & 4.97  & 6.28  & \textbf{6.57} & 5.41  & \textbf{5.75}   \\
    \hline
    \hline
    \multirow{8}[0]{*}{0.03} & Noisy &8.54  & 10.47 & 10.61 & 11.45 & 9.74  & 11.59 & 8.58  & 11.83 & 10.63 & 9.57  & 10.30 \\
\cline{2-13}          & APSS  & 8.60  & 9.17  & 7.98  & 8.92  & 7.80  & 6.96  & 10.52 & 8.29  & 9.37  & 8.25  & 8.59   \\
\cline{2-13}          & RIMLS & 9.66  & 10.06 & 9.59  & 10.87 & 10.00 & 8.66  & 9.83  & 9.62  & 10.34 & 10.15 & 9.88   \\
\cline{2-13}          & AWLOP & 7.15  & 9.38  & 9.76  & 10.62 & 7.86  & 10.52 & 6.53  & 11.15 & 9.50  & 8.24  & 9.07   \\
\cline{2-13}          & NLD   & 8.03  & 9.65  & 9.55  & 10.10 & 9.02  & 10.20 & 8.25  & 10.44 & 9.57  & 8.94  & 9.37   \\
\cline{2-13}          & MRPCA & 8.13  & 9.61  & 9.50  & 9.91  & 8.97  & 9.78  & 8.33  & 10.18 & 9.70  & 9.01  & 9.31   \\
\cline{2-13}          & LR    & \textbf{5.93} & 7.39  & \textbf{6.69} & 7.77  & 6.45  & \textbf{6.25} & 6.83  & \textbf{7.72} & 7.76  & \textbf{6.37} & 6.92   \\
\cline{2-13}          & GLR   & 5.96  & \textbf{7.34} & 6.90  & \textbf{7.75} & \textbf{6.28} & 6.71  & \textbf{6.07} & 7.76  & \textbf{7.48} & 6.48  & \textbf{6.87}   \\
    \hline
    \hline
    \multirow{8}[0]{*}{0.04} & Noisy & 10.23 & 12.32 & 12.53 & 13.47 & 11.98 & 13.93 & 10.78 & 13.89 & 12.61 & 11.45 & 12.32  \\
\cline{2-13}          & APSS  & 11.87 & 11.46 & 10.30 & 12.33 & 11.24 & 9.48  & 12.88 & 10.92 & 12.01 & 12.51 & 11.50   \\
\cline{2-13}          & RIMLS & 11.92 & 13.36 & 12.25 & 13.95 & 12.89 & 11.78 & 12.68 & 13.62 & 13.23 & 12.88 & 12.86   \\
\cline{2-13}          & AWLOP & 9.01  & 11.55 & 11.95 & 12.96 & 10.40 & 13.25 & 8.68  & 13.52 & 11.76 & 10.47 & 11.35   \\
\cline{2-13}          & NLD   & 9.83  & 11.71 & 11.81 & 12.52 & 11.42 & 12.96 & 10.48 & 12.94 & 11.81 & 11.01 & 11.65   \\
\cline{2-13}          & MRPCA & 9.94  & 11.70 & 11.81 & 12.39 & 11.38 & 12.62 & 10.56 & 12.78 & 11.88 & 11.08 & 11.61   \\
\cline{2-13}          & LR    & 7.87  & \textbf{8.60} & 7.86  & \textbf{8.98} & 8.70  & \textbf{7.88} & 9.64  & \textbf{8.93} & 8.65  & 8.41  & 8.55   \\
\cline{2-13}          & GLR   & \textbf{7.25} & 8.79  & \textbf{8.43} & 9.29  & \textbf{7.88} & 8.31  & \textbf{7.20} & 9.40  & \textbf{8.38} & \textbf{7.96} & \textbf{8.29}   \\
    \hline
    \end{tabular}%
  \label{tab:shapenetmcd}%
\end{table*}%

\section{Conclusion}
\label{sec:con}
In this paper, we propose a graph Laplacian regularization based 3D point cloud denoising algorithm. To utilize the self-similarity among surface patches, we adopt the low-dimensional manifold prior, and collaboratively denoise the patches by minimizing the manifold dimension. 
To compute manifold dimension with discrete patch observations, we approximate the manifold dimension with a graph Laplacian regularizer, and construct the patch graph with a new measure for the discrete patch distance. 
The proposed scheme is shown to have graph spectral low-pass filtering interpretation and numerical stability in solving the linear equation system, and efficient to solve with methods like PCG. 
Experimental results suggest that our proposal outperforms existing schemes with better structural detail preservation.

\appendices

\section{}
\label{sec:append}
Assume that $\mathcal{M}$ is a Riemannian manifold with boundary, equipped with the probability density function (PDF) $h(\mathbf{p})$ describing the distribution of the vertices on $\mathcal{M}$,
and that $\boldsymbol{\alpha}_i$ belongs to the class of $\kappa$-H$\ddot{o}$lder functions \cite{hein2006uniform} on $\mathcal{M}$. Then according to the proof in \cite{pang2017graph}, there exists a constant $c$ depending only on $C_r$ such that for $\kappa \ge 3$ and the weight parameter $\epsilon = O\left( M^{-\frac{\kappa}{2\kappa+2\delta+\delta^2+\delta\kappa}} \right)$, where $\delta$ denotes the manifold dimension, such that\footnote{We refer readers to \cite{hein2006uniform} for the uniform convergence result in a more general setting and its corresponding assumptions on $\mathcal{M}$, $\epsilon$, and the graph weight kernel function $\psi(\cdot)$.}

\begin{multline} \label{eq:hein1}
\mathrm{sup} \left| \frac{cM^{2\gamma-1}}{\epsilon^{4(1-\gamma)}(M-1)}S_{\mathbf{L}}(\boldsymbol{\alpha}_i) - S_{\Delta}(\alpha_i)  \right| \\
= O\left( M^{-\frac{\kappa}{2\kappa+2\delta+\delta^2+\delta\kappa}} \right), 
\end{multline}
where $S_{\Delta}(\alpha_i)$ is induced by the $2(1-\gamma)$-th weighted Laplace-Beltrami operator on $\mathcal{M}$, which is given as
\begin{equation} 
S_{\Delta}(\alpha_i) = \int_{\mathcal{M}} \|\nabla_{\mathcal{M}} \alpha_i(\mathbf{p})\|_2^2 h(\mathbf{p})^{2(1-\gamma)} d \mathbf{p}.
\end{equation}

Assuming that the vertices are uniformly distributed on $\mathcal{M}$, then 
\begin{equation}
\int_{\mathcal{M}} h(\mathbf{p}) d \mathbf{p} = 1,\quad h(\mathbf{p}) = \frac{1}{|\mathcal{M}|},
\end{equation}
where $|\mathcal{M}|$ is the volume of the manifold $\mathcal{M}$.
For implementation, similar to the setting in \cite{pang2017graph}, we set $\gamma=0.5$, then $S_{\Delta}(\alpha_i)$ becomes 
\begin{equation} \label{eq:hein2}
S_{\Delta}(\alpha_i) = \frac{1}{|\mathcal{M}|}\int_{\mathcal{M}} \|\nabla_{\mathcal{M}} \alpha_i(\mathbf{p})\|_2^2 d \mathbf{p}.
\end{equation}
From (\ref{eq:hein1}) and (\ref{eq:hein2}), the convergence in (\ref{eq:approx0}) is readily obtained by weakening the uniform convergence of (\ref{eq:hein1}) to point-wise convergence.
% \begin{equation} 
% \underset{M \rightarrow \infty, \epsilon \rightarrow 0} {\lim} |\mathcal{M}| \sum_{i=1}^{3k} \boldsymbol{\alpha}_i^\top \mathbf{L} \boldsymbol{\alpha}_i
% \sim \sum_{i=1}^{3k} \int_{\mathcal{M}} \|\nabla_{\mathcal{M}} \alpha_i(\mathbf{p})\|_2^2 d \mathbf{p},
% \end{equation}
% where $\sim$ means there exists a constant depending on $M$, $C_r$ and $\gamma$, such that the equality holds.
% The rate of convergence $O\left( M^{-\frac{\kappa}{2\kappa+2\delta+\delta^2+\delta\kappa}} \right)$ depends on $M$ and $\delta$.
% As the number of samples $M$ increases and the neighborhood size $r=\epsilon C_r$ shrinks, the graph Laplacian regularization approaches its continuous limit.
% Moreover, if the manifold dimension $\delta$ is low, we can ensure a good approximation of the continuous regularization functional even if the manifold is embedded in a high-dimensional space.
% Consequently, given a point cloud, one can approximate the dimension of $\mathcal{M}$ with the $\boldsymbol{\alpha}_i$'s and the constructed graph Laplacian $\mathbf{L}$. 

\ifCLASSOPTIONcaptionsoff
  \newpage
\fi

\bibliographystyle{IEEEtran}
% \bibliography{./refs}
% Generated by IEEEtran.bst, version: 1.12 (2007/01/11)

\end{document}